\documentclass[journal]{IEEEtran}
\usepackage{bm}
\usepackage{amsfonts}
\usepackage{amsmath}
\usepackage{booktabs}
\usepackage{multirow}
\usepackage{cite}
\usepackage{array}
\usepackage{pifont}
\usepackage{tikz}
\usepackage{ulem}
\usepackage{hyperref}
\usepackage{subfigure}
\PassOptionsToPackage{table}{xcolor}
\usepackage{xcolor}

\begin{document}
\title{Video-Level Language-Driven Video-Based Visible-Infrared Person Re-Identification}
\author{Shuang~Li, Jiaxu~Leng, Changjiang~Kuang, Mingpi~Tan and Xinbo Gao, \IEEEmembership{Fellow, IEEE} % <-this % stops a space

\thanks{This work was supported in part by the National Key R\&D Program of China  under Grant No.2022YFA1004100, in part by the National Natural Science Foundation of China under Grants No. 62472060, U23A20318, and 62221005, in part by the Natural Science Foundation of Chongqing under Grant    No. CSTB2024NSCQ-QCXMX0060, CSTB2023NSCQ-LZX0061, and CSTB2022NSCQ-MSX0547, in part by the Science and Technology Research Program of Chongqing Municipal Education Commission under Grant No. KJZD-K202300604, KJQN202400648, in part by the China Postdoctoral Science Foundation under Grants No. GZC20233362, 2024MD754043, in part by the Chongqing Institute for Brain and Intelligence, in part by Chongqing University of Postsand Telecommunications Ph.D. Innovative Talents Project underGrants BYJS202401. (Corresponding author: J. Leng, X. Gao.)}
\thanks{S. Li, J. Leng, C. Kuang, M. Tan and X. Gao,  are with the School of Computer Science and Technology, Chongqing University of Posts and Telecommunications, Chongqing, China (E-mail: shuangli936@gmail.com, lengjx@cqupt.edu.cn, 230201049@stu.cqupt.edu.cn, tanmingp11@163.com,  gaoxb@cqupt.edu.cn). They are also affiliated with the Chongqing Institute for Brain and Intelligence, Guangyang Bay Laboratory, Chongqing, China.}
}

%\markboth{Please submit the manuscript to the Special Issue on Deep Learning for Intelligent Media Computing and Applications}%
%{Shell \MakeLowercase{\textit{et al.}}}
\markboth{IEEE Transactions on Information Forensics and Security}%
{Shell \MakeLowercase{\textit{et al.}}: Bare Advanced Demo of IEEEtran.cls for IEEE Computer Society Journals}
\maketitle

\begin{abstract}
Video-based Visible-Infrared Person Re-Identification (VVI-ReID) aims to match pedestrian sequences across modalities by extracting modality-invariant sequence-level features. As a high-level semantic representation, language provides a consistent description of pedestrian characteristics in both infrared and visible modalities. Leveraging the Contrastive Language-Image Pre-training (CLIP) model to generate video-level language prompts and guide the learning of modality-invariant sequence-level features is theoretically feasible.
%Editor 3: Please note that ``CLIP'' has not been defined. Abbreviations and acronyms are often defined the first time the term is used within the abstract and again in the main text and then used throughout the remainder of the document. Please consider adhering to this convention. The target journal may have a list of abbreviations that are considered common enough that they do not need to be defined.
However, the challenge of generating and utilizing modality-shared video-level language prompts to address modality gaps remains a critical problem.
To address this problem, we propose a simple yet powerful framework, video-level language-driven VVI-ReID (VLD), which consists of two core modules: invariant-modality language prompting (IMLP) and spatial-temporal prompting (STP).  
IMLP employs a joint fine-tuning strategy for the visual encoder and the prompt learner to effectively generate modality-shared text prompts and align them with visual features from different modalities in CLIP's multimodal space, thereby mitigating modality differences.  
Additionally, STP models spatiotemporal information through two submodules, the spatial-temporal hub (STH) and spatial-temporal aggregation (STA), which further enhance IMLP by incorporating spatiotemporal information into text prompts. The STH aggregates and diffuses spatiotemporal information into the [CLS] token of each frame across the vision transformer (ViT) layers, whereas STA introduces dedicated identity-level loss and specialized multihead attention to ensure that the STH focuses on identity-relevant spatiotemporal feature aggregation.  
The VLD framework achieves state-of-the-art results on two VVI-ReID benchmarks. On the HITSZ-VCM dataset, it improves the Rank-1 accuracy by 7.3\% and mAP by 7.6\% (infrared-to-visible) and the Rank-1 accuracy by 10.4\% and the mAP accuracy by 9.3\% (visible to infrared) and requires only 2 hours of training, 2.39M additional parameters, and 0.12G FLOPs. 
The code will be released at \url{https://github.com/Visuang/VLD}.

\end{abstract}
\IEEEpeerreviewmaketitle
\section{Introduction}
\IEEEPARstart{V}{ideo-based} person re-identification (ReID) aims to match individuals across different camera views by utilizing the rich spatiotemporal information embedded in video sequences \cite{ gu2020appearance, liu2021watching, wang2021pyramid, bai2022salient, liu2023deeply, liu2024TMT, yu2023tf,wang2022body}. Significant advancements in this field have been driven by breakthroughs in deep learning \cite{ye2021deep,zheng2015partial,li2018harmonious,leng2019survey, ye2021dynamic, yang2023dual, ye2024securereid} and the development of large-scale annotated datasets \cite{ristani2016performance,wei2018person,zheng2016mars,li2019GLTR,wu2018exploit,wang2014person,hirzer2011person}. 
Despite this progress, achieving robust performance in 24-hour surveillance systems remains a formidable challenge because of the spectral differences between daytime and nighttime imagery. While daytime video sequences are captured with RGB cameras, nighttime sequences are captured with near-infrared cameras, resulting in a substantial modality gap that traditional video-based ReID methods struggle to overcome. 
To address this limitation, VVI-ReID \cite{lin2022learning} has emerged as a promising method. By enabling cross-modality matching, VVI-ReID bridges the gap between the visible and infrared modalities, offering significant practical value for day-and-night surveillance systems and enhancing their reliability under varying lighting conditions.

\begin{figure}[t!]
\centering
\includegraphics[width=8.9cm,keepaspectratio=true]{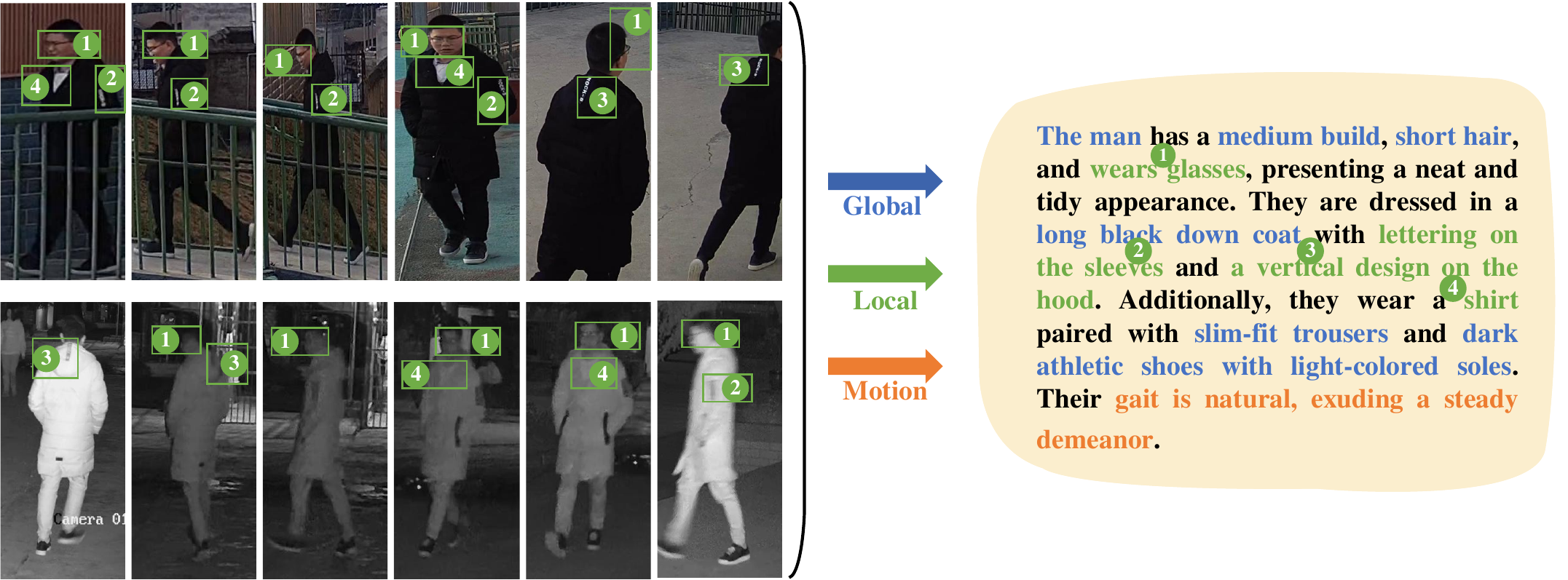}
\caption{Modality-shared video-level language effectively describes pedestrian sequences across different modalities, integrating global information (blue), local information (green), and motion information (orange). The numbered markers indicate frame-specific local features. }
\label{intro1}
\end{figure}

To alleviate modality differences, researchers have proposed two main methods: feature-level methods and image-level methods. Feature-level methods use adversarial learning to embed features from different modalities into a unified feature space \cite{lin2022learning,li2023adversarial, zhou2023video}, whereas image-level methods leverage Generative Adversarial Networks (GANs) \cite{du2023video} or edge detection operators \cite{li2023intermediary} to generate intermediate modality images, helping the model focus on shared features.
Despite these advancements, the performance of these methods remains constrained by the coarse-grained supervision of traditional one-hot labels, which fail to provide the fine-grained guidance necessary for effective cross-modal alignment.
In contrast, as shown in Fig. \ref{intro1}, language descriptions offer richer supervision: 1) \textbf{language provides a shared semantic representation across modalities}, acting as an intermediate bridge to alleviate modality differences; and 2) \textbf{video-level language descriptions capture global, local, and motion information}, enhancing the integration of complementary features.

\begin{figure}[t!]
\centering
\includegraphics[width=8.9cm,keepaspectratio=true]{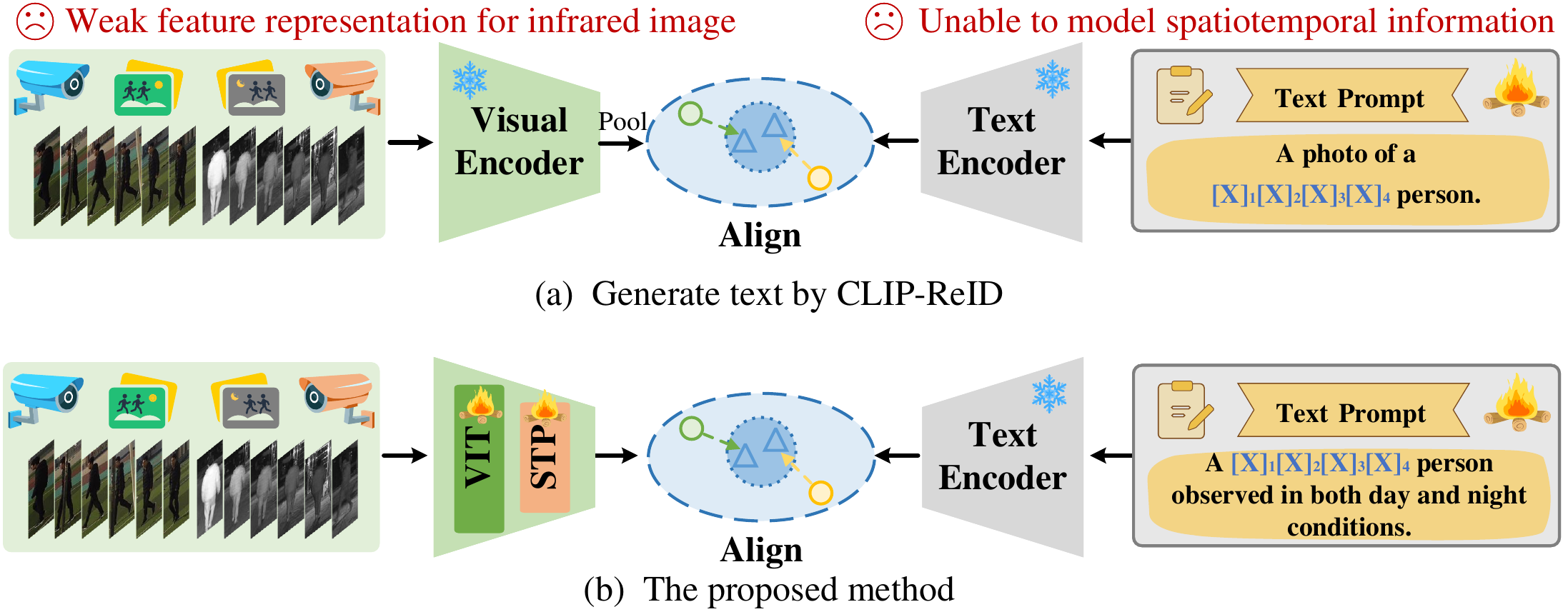}
\caption{The core motivations of this paper:  
(a) CLIP-ReID struggles to model the spatiotemporal information of pedestrian sequences and exhibits weak feature representation for the infrared image.  
(b) To overcome these limitations, we propose a novel framework that leverages CLIP to generate video-level language prompts as a bridge to mitigate modality differences. This is achieved through the joint fine-tuning of the visual encoder and text prompt learner using the proposed IMLP and STP.  }
\label{intro2}
\end{figure}

However, owing to the high cost of annotation, pedestrian sequences often lack corresponding modality-shared video-level text descriptions. Recently, CLIP-ReID \cite{li2023clip}, as illustrated in Fig.\ref{intro2} (a), emerged as a pioneer in the ReID field by exploring the use of the multimodal knowledge of Contrastive Language-image Pre-training (CLIP) to learn text descriptions for pedestrian images, effectively addressing the issue of missing textual annotations. Inspired by this work, we attempt to extend this learning method directly to VVI-ReID. However, this adaptation did not yield satisfactory results, primarily because of two inherent limitations of the original CLIP model: (1) \textbf{CLIP's training data consist predominantly of RGB images}, which constrains its ability to represent infrared features, resulting in text prompts that lack modality-shared information; and (2) \textbf{CLIP is pretrained on single-frame images}, so it cannot model spatiotemporal information, leading to text prompts without spatiotemporal features.

To address these limitations, this paper proposes a novel framework, \textbf{v}ideo-level \textbf{l}anguage-\textbf{d}riven VVI-ReID (\textbf{VLD}), as illustrated in Fig.\ref{intro2} (b). This framework simultaneously fine-tunes the vision encoder and the text prompt learner, overcoming the limitations of the original CLIP model in handling multispectral data and enabling it to generate modality-shared text prompts.
Revisiting CLIP-ReID, its first stage can be viewed as generating text prototypes, whereas the second stage constructs a classifier based on these prototypes and fine-tunes the vision encoder. This process resembles traditional image classification tasks, where the image encoder and classifier are jointly optimized. Inspired by this design, we propose \textbf{invariant-modality language prompting} (\textbf{IMLP}), which achieves deep alignment between text and visual features through joint fine-tuning of the vision encoder and the text prompt learner. Specifically, we treat the combination of the frozen text encoder and the text prompt learner as a learnable classifier, where class labels correspond to identity labels and classifier prototypes are derived from text features generated by the text encoder. This method enhances the vision encoder's ability to represent different modalities and generates modality-shared text prompts aligned with pedestrian identities, effectively reducing the modality gap.

Moreover, owing to the lack of spatiotemporal modeling capabilities in the original CLIP model, the text prompts generated by IMLP fail to incorporate spatiotemporal information. A straightforward solution to this problem is to introduce spatiotemporal modeling mechanisms into the vision encoder \cite{li2023intermediary,lin2022learning,feng2024cross,zhou2023video,yu2023tf,deng2023prompt}.  
The existing VVI-ReID methods typically add LSTM \cite{li2023intermediary,lin2022learning}, transformer \cite{feng2024cross}, or GCN \cite{zhou2023video} networks after the vision encoder to enhance spatiotemporal interactions. Similarly, the video-based ReID method TF-CLIP \cite{yu2023tf} allows frame-level memories within a sequence to communicate, extracting temporal information from interframe relations and extending the spatiotemporal modeling capabilities of the ViT. However, these methods overlook spatiotemporal interactions within the intermediate layers of the vision encoder. This oversight prevents the model from fully leveraging the interactions between global and local spatial information and temporal dynamics in intermediate layers, thereby limiting its ability to model complex spatiotemporal relationships. Additionally, these methods significantly increase the computational overhead.

To address this issue, we propose a lightweight \textbf{spatial-temporal prompting} (\textbf{STP}) mechanism, which requires only \textbf{2.39M} parameters and \textbf{0.12G} FLOPS. STP enables layer-by-layer interframe interactions within the ViT and integrates these interactions into sequence-level feature representations. Specifically, we design a \textbf{spatial-temporal hub} (\textbf{STH}), a learnable tensor spanning temporal, spatial, and channel dimensions, which serves as the central hub for spatiotemporal information exchange. In the \(i\)-th layer, the STH is concatenated with patch tokens along the spatial dimension, allowing spatial information to be aggregated via multihead attention. In the \(i+1\)-th layer, the transposed STH is concatenated with patch tokens again, facilitating spatiotemporal interactions and diffusing the aggregated information to the [CLS] token of each frame.
Furthermore, we propose a \textbf{spatial-temporal aggregation} (\textbf{STA}) mechanism to ensure that the STH effectively aggregates identity-specific spatiotemporal information. STA incorporates a specially designed multihead attention mechanism to establish a direct connection between the [CLS] token and the STH. Additionally, an identity-level loss is introduced to enhance the ability of the STH to aggregate and represent spatiotemporal information effectively.

 Our main contributions are summarized as follows:
\begin{itemize}
    \item We propose VLD, a novel framework for learning modality-invariant sequence-level pedestrian representations, guided by video-level language prompts.
    \item We propose invariant-modality language prompting (IMLP), which effectively generates and utilizes modality-shared language prompts to mitigate modality differences.
    \item We propose spatial-temporal prompting (STP), a lightweight mechanism for the layer-by-layer aggregation and diffusion of spatiotemporal information within the ViT, enabling efficient interframe interaction.
    \item Experiments on the HITSZ-VCM and BUPTCampus datasets demonstrate that VLD achieves state-of-the-art performance while requiring only 2 hours of training, an additional 2.39M parameters, and 0.12G FLOPs.
\end{itemize}

The rest of this paper is organized as follows. Section \uppercase\expandafter{\romannumeral2} introduces related work; Section \uppercase\expandafter{\romannumeral3}  elaborates the proposed method; Section \uppercase\expandafter{\romannumeral4} analyzes the comparative experimental results; and Section \uppercase\expandafter{\romannumeral5} concludes this paper.

\section{Related Work}
\subsection{Image-based Visible-Infrared Person Re-Identification}
Person re-identification (ReID) \cite{wang2025idea,wang2025decoupled,wang2025mambapro,li2023logical} retrieves images of the same identity across cameras. 
Traditional single-modality ReID methods have made significant progress. To enhance generalization, GLAVF \cite{li2020attribute} effectively converts semantic attributes into domain-invariant visual proxies, providing a robust prior for transferable representation learning. Meanwhile, HCM \cite{si2022hybrid} presents a representative advancement by jointly leveraging identity- and image-level contrastive constraints to effectively mine hard pedestrian samples, thereby enhancing performance in an unsupervised setting.
Despite these advances in general ReID \cite{wang2025idea,wang2025decoupled,wang2025mambapro,si2022hybrid,li2020attribute}, VI-ReID remains challenging due to significant intermodality discrepancies. Existing image-based VI-ReID methods can be broadly categorized into image-level\cite{wang2019learning,si2023tri,li2020infrared,choi2020hi,wei2021syncretic} and feature-level alignment \cite{wu2021discover,liu2022learning,zhang2023mrcn,zhang2023diverse,cheng2023cross,si2023diversity} strategies.

Image-level modality alignment addresses modality differences through style transfer. D2RL \cite{wang2019learning} combines image- and feature-level alignment by reducing modality differences via style transfer and minimizing appearance gaps through feature embedding.
TCOM \cite{si2023tri} introduces a tri-modality consistency optimization framework that generates heterogeneous augmented images to bridge modality gaps and enhance feature alignment, demonstrating strong capability in modeling cross-modality relationships.
XIV \cite{li2020infrared} introduces an auxiliary X modality to bridge the visible-infrared gap. Hi-CMD \cite{choi2020hi} employs hierarchical modality disentanglement to remove nonidentity information, such as pose and illumination, ensuring identity-discriminative features. SMCL \cite{wei2021syncretic} generates a synthetic modality by merging visible and infrared features, using homogeneity and distributional similarity learning to map heterogeneous features into a unified space, thereby enhancing cross-modality discrimination.  

Feature-level modality alignment projects visible and infrared images into a unified feature space to mitigate discrepancies. MPANet \cite{wu2021discover} combines modality alleviation and pattern alignment to extract fine-grained features for improved discrimination. MAUM \cite{liu2022learning} incorporates unidirectional metrics and a memory bank to strengthen feature associations, particularly in imbalanced modality scenarios. MRCN \cite{zhang2023mrcn} employs instance normalization to disentangle modality-relevant and irrelevant features, refining alignment via restitution and compensation modules. DEEN \cite{zhang2023diverse} generates diverse embeddings for robust cross-modality feature learning under low-light conditions, enhancing real-world applicability. 
DFC \cite{si2023diversity} unifies clustering and instance-level constraints to align cross-modality features without annotations, achieving strong performance in unsupervised VI-ReID.
However, these methods overlook the rich spatiotemporal information in video sequences, which limits their practical application.

\subsection{Video-based Visible-Infrared Person Re-Identification}
Recent advancements in VVI-ReID have introduced innovative methods to enhance cross-modality feature alignment and leverage temporal information for improved matching accuracy. MITML \cite{lin2022learning} addresses challenges in visible-infrared video matching by learning modality-invariant and motion-invariant features, effectively mitigating modality discrepancies. 
IBAN \cite{li2023intermediary} introduces an elegant framework that jointly addresses modality discrepancy and temporal noise in video-based VI-ReID, achieving strong performance via intermediary-guided alignment and bidirectional aggregation.
AuxNet \cite{du2023video} introduces auxiliary samples to augment video sequence learning by combining primary and auxiliary data. It employs GANs for modality alignment and utilizes curriculum learning to effectively exploit temporal information from trajectories, significantly improving the model’s performance. SAADG \cite{zhou2023video} addresses intra- and intermodality discrepancies through style perturbation attacks and captures multiview and cross-modality associations using a graph-based dual interaction module.  
CST \cite{feng2024cross} employs a transformer-based architecture to capture both spatial and long-range temporal dependencies in ultra-long video sequences, marking a shift from traditional CNN or RNN models to global temporal modeling.
Despite recent progress in VVI-ReID through adversarial, graph-based, and transformer-based models, most existing methods overlook the potential of leveraging language to enhance fine-grained, sequence-level cross-modal representations.

\subsection{Prompt Learning and Vision-Language Models in Re-Identification}
Prompt learning has emerged as a parameter-efficient strategy for adapting large-scale vision-language models such as CLIP \cite{radford2021learning} to downstream tasks. By introducing learnable or handcrafted prompts, it enables task-specific adaptation without tuning the backbone, showing strong generalizability across classification, segmentation, and retrieval \cite{zhou2023zegclip, ma2022x, tang2021clip4caption}.
Recently, the integration of prompt learning and vision-language models has been extended to a range of applications. GSNet\cite{ye2025towards} utilizes prompt learning to guide feature fusion, effectively combining domain-specific knowledge with general vision-language representations for open-vocabulary remote sensing image segmentation. MugTracker\cite{zhu2024multi} incorporates multimodal understanding and image-to-text generation to update target descriptions dynamically, thereby mitigating inconsistencies between visual and linguistic representations and significantly enhancing the robustness and accuracy of vision-language tracking (VLT). 
CPIPTrack\cite{zhu2024vision} introduces three types of interactive prompts to fully activate CLIP's cross-modal potential, achieving lightweight and efficient fusion of visual and language features for improved tracking performance.

In the field of ReID, CLIP-based models have been actively explored to address challenges such as domain shifts, modality gaps, and semantic alignment \cite{yan2023clip,jiang2023cross, li2023clip, he2023region, yu2023tf,yu2025climb,yu2024clip}. For text-to-image person retrieval (TI-ReID) \cite{MANet}, CFine \cite{yan2023clip} leverages CLIP's image encoder for multigranularity feature learning, enhancing fine-grained feature alignment, whereas IRRA \cite{jiang2023cross} employs both image and text encoders to achieve cross-modal local feature alignment via a dual-encoder architecture.
For image-based ReID lacking textual annotations, CLIP-ReID \cite{li2023clip} proposes a two-stage framework that learns identity-specific textual prompts to guide visual feature learning. Beyond conventional paired training paradigms, U-DG \cite{li2025breaking} introduces a novel prompt-based framework that constructs identity- and perturbation-guided textual prompts from unpaired data, enabling strong domain generalization without relying on cross-camera identity correspondences.
In video-based ReID tasks, TF-CLIP \cite{yu2023tf} bypasses the text encoder by storing identity-specific CLIP memory vectors, enabling text-free feature alignment. To further improve efficiency and spatiotemporal modeling, CLIMB-ReID \cite{yu2025climb} introduces a hybrid CLIP-Mamba architecture that achieves efficient CLIP knowledge transfer and multi-scale temporal modeling for superior person ReID performance.
In VI-ReID, where modality discrepancy is more severe, CSDN \cite{yu2024clip} adopts a three-stage framework to generate textual descriptions for each modality, leveraging CLIP's semantic space for cross-modal alignment via language supervision.

Building on these advances, this paper extends CLIP to VVI-ReID by proposing the VLD method, which captures spatiotemporal semantics across frames and aligns cross-modal features within CLIP's multimodal embedding space.

\section{Proposed method}
In this section, we present the proposed VLD framework, as illustrated in Fig.\ref{framework}. We begin with a brief review of CLIP and introduce a CLIP-based VVI-ReID baseline method. Then, we provide a detailed explanation of the proposed VLD framework, including its two modules, invariant-modality language prompting (IMLP) and spatial-temporal prompting (STP), as well as the training and testing processes.

\subsection{Brief Review of CLIP}\label{sec3.1}
Contrastive Language-Image Pre-training (CLIP) is trained on a large-scale dataset of image-text pairs, utilizing a vision encoder and a text encoder to extract features from images and text, respectively. By aligning image features with corresponding text features through a contrastive loss function, CLIP achieves precise cross-modal matching, making it highly suitable for diverse downstream tasks.
In the field of ReID, CLIP exhibits strong potential by leveraging frozen vision and text encoders—together with a prompt learner—to generate identity-level textual representations for cross-modal alignment. However, as CLIP is designed primarily for image-text alignment tasks and trained on datasets dominated by visible images, it lacks the ability to model spatiotemporal information and struggles to represent infrared pedestrian images. These limitations make CLIP unsuitable for direct application in video-based VVI-ReID.

\subsection{CLIP-driven Baseline for VVI-ReID}\label{sec3.2}
\begin{figure*}[t!]
\centering
\includegraphics[width=18cm,keepaspectratio=true]{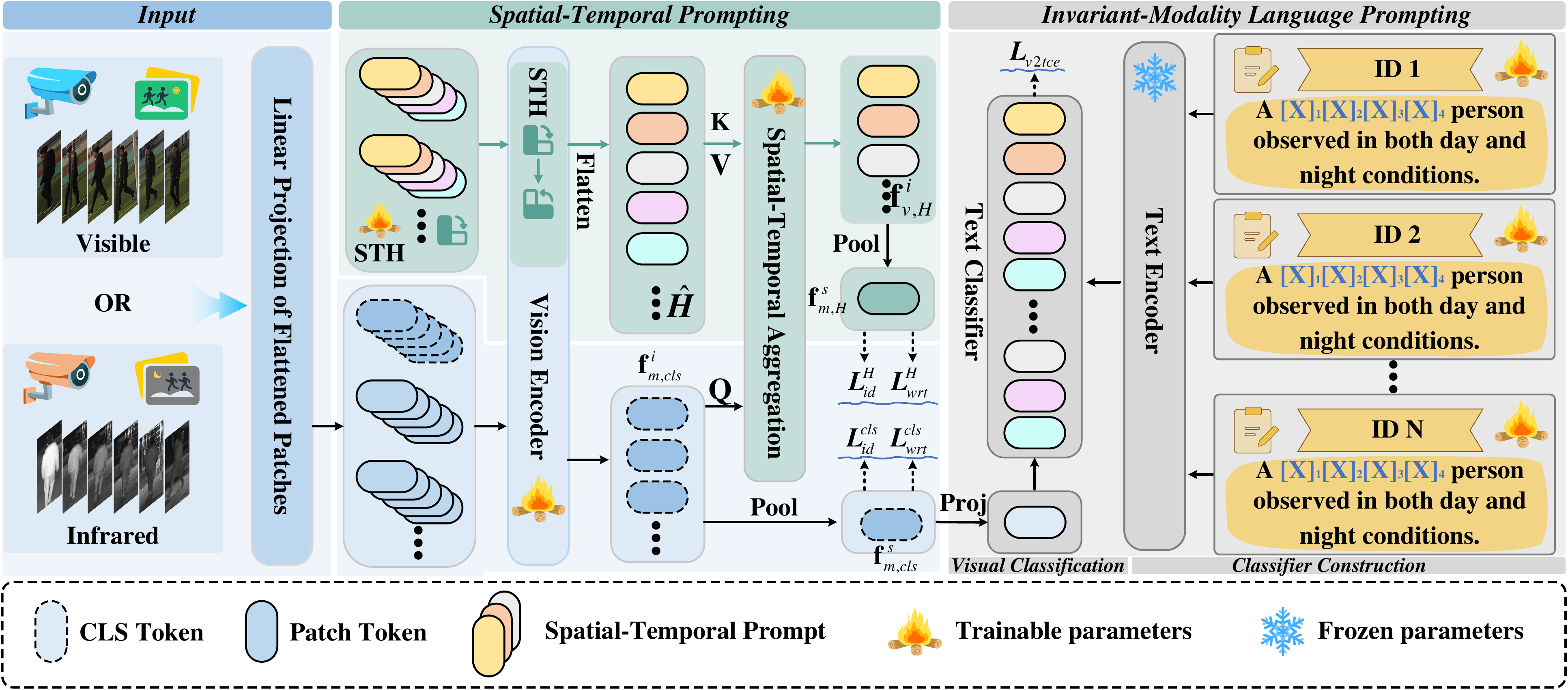}
\caption{The proposed VLD framework consists of two modules: invariant-modality language prompting (IMLP) and spatial-temporal prompting (STP). IMLP treats the frozen text encoder and the prompt learner as an identity classifier while fine-tuning the vision encoder and the prompt learner to align language with different visual modalities, thus reducing modality discrepancies. STP provides CLIP with spatiotemporal modeling capabilities, where the STH in STP acts as a central hub, aggregating and diffusing spatiotemporal information to the [CLS] tokens of different frames, with STA further enhancing the spatiotemporal aggregation ability of the STH.
}
\label{framework}
\end{figure*}

As shown in Fig. \ref{framework}, the input to VLD consists of visible and infrared pedestrian sequences, denoted as \( V = \{V^{t} \mid V^t \in \mathbb{R}^{H \times W} \}_{t=1}^{T} \) and \( I = \{I^{t} \mid I^t \in \mathbb{R}^{H \times W} \}_{t=1}^{T} \), where \( T \) represents the total number of images in the sequence.
Following the standard procedure in VVI-ReID\cite{lin2022learning}, we use the visual encoder of CLIP to extract features from each frame in the pedestrian sequence. Then, temporal average pooling is applied to the extracted features to generate a sequence-level representation.
Specifically, taking the visible sequence \( V \) as an example, consider each image \( V^{t} \in \mathbb{R}^{H \times W \times C} \) in the sequence, where \( H \) and \( W \) denote the height and width of the \( t \)-th frame, respectively. The image is first divided into \( N = H \times W / P^2 \) nonoverlapping image patches \( \textbf{f}_{vis}^{\hspace{0.1em} t, i} \in \mathbb{R}^{3 \times P^2} \). These patches are then mapped to a 1D token sequence through a linear projection matrix \( \textbf{C} \in \mathbb{R}^{3P^2 \times D} \), and spatial position embeddings \( \text{e}^{spa} \) are added to these tokens, where $D$ represents the feature dimension.
Additionally, a [CLS] token is prepended to the patch sequence to summarize the global features of the input image. 
% The input to the visual encoder for frame \( t \), represented as \( \textbf{f}_{vis}^{\hspace{0.1em} t} \), 
This process is formulated as:  
\begin{equation}
\begin{aligned}
\textbf{\text{f}}_{vis}^{\hspace{0.1em} t} =[\textbf{\text{f}}_{vis,cls}^{\hspace{0.1em} t}, \textbf{C}\textbf{\text{f}}_{vis}^{\hspace{0.1em} t,1}, \dots, \textbf{C}\textbf{\text{f}}_{vis}^{\hspace{0.1em} t,N}] + \text{e}^{spa}
\end{aligned},
\end{equation}
Then, we input $\textbf{f}_{vis}^{\hspace{0.1em} t}$ into the visual encoder $E_{v}$ to obtain the feature representation $\textbf{f}_{vis,cls}^{\hspace{0.1em} t}$ for the image. Finally, temporal average pooling $\text{AvgPool}$ is applied to all frame representations to obtain the sequence-level feature representation:
\begin{equation}
\begin{aligned}
\textbf{\text{f}}_{vis,cls}^{\hspace{0.1em} s} =\text{AvgPool}([\textbf{\text{f}}_{vis,cls}^{\hspace{0.1em} 1},\textbf{\text{f}}_{vis,cls}^{\hspace{0.1em} 2},\dots,\textbf{\text{f}}_{vis,cls}^{\hspace{0.1em} T}])
\end{aligned},
\end{equation}
To ensure that the sequence-level visible feature $\textbf{f}_{vis,cls}^{\hspace{0.1em} s}$ and infrared pedestrian feature $\textbf{f}_{ir,cls}^{\hspace{0.1em} s}$ possess strong identity discriminability across modalities, we follow AGW \cite{ye2021deep}, employing both cross-entropy loss $\bm L_{id}^{cls}$ and weighted regularized triplet loss $\bm L_{wrt}^{cls}$:
\begin{equation}
\begin{aligned}
\bm L_{id}^{cls}&=-\frac{1}{n_{b}}\sum_{i=1}^{n_{b}}q_{i}\log(\bm W_{id}(\textbf{f}_{m,cls}^{\hspace{0.1em} i,s}))
\end{aligned},
\end{equation}
\begin{equation}
\begin{aligned}
% \begin{split}
\bm L_{wrt}^{cls}= \frac{1}{n_{b}}\sum_{i=1}^{n_{b}}\log(1+\exp(\sum_{i,j} w_{i,j}^{p}\bm d_{i,j}^{p}-\sum_{i,k} w_{i,k}^{n}\bm d_{i,k}^{n}),\\
w_{i,j}^{p} = \frac{\exp(\bm d_{i,j}^{p})}{\sum_{\bm d_{i,j}\in{P_{i}}}\exp(\bm d_{i,j}^{p})},w_{i,k}^{n} = \frac{\exp(-\bm d_{i,k}^{n})}{\sum_{\bm d_{i,k}\in{N_{i}}}\exp(-\bm d_{i,k}^{n})}
% \end{split}
\end{aligned},
\end{equation}
where $\textbf{f}_{m,\text{cls}}^{\hspace{0.1em} i,s}$ represents the $i$-th sequence-level feature in a batch and $m$ denotes the modality (visible $vis$ or infrared $ir$). The indices $j$ and $k$ (i.e., $P_{i}$ and $N_{i}$) correspond to the positive and negative samples (or sample sets) associated with the anchor sample within the batch, respectively. The Euclidean distance between two features is denoted as $\bm{d}_{i,j}$ and is computed as:
$\bm{d}_{i,j} = \|\textbf{f}_{m,\text{cls}}^{\hspace{0.1em} i,s} - \textbf{f}_{m,\text{cls}}^{\hspace{0.1em} j,s}\|_{2}$.

\subsection{Invariant-Modality Language Prompting}\label{sec3.3}
Although the model architecture based on the CLIP visual encoder demonstrates decent performance, its potential remains significantly constrained due to the lack of effective incorporation of text prompt guidance. This limitation is reflected in two main aspects: \textbf{1) the absence of text prompts as a bridge hinders the model's ability to mitigate discrepancies between different modalities};  \textbf{2) fine-tuning only the unimodal visual encoder risks disrupting CLIP's original multimodal embedding space, which limits the full utilization of its rich multimodal knowledge}.
To address these issues, we propose invariant-modality language prompting (IMLP), inspired by the classical image classification pipeline, which jointly trains a feature extractor and a classifier. In IMLP, the CLIP visual encoder acts as the feature extractor, while learnable identity-level text prompts and a frozen text encoder serve as the classifier. IMLP comprises two key processes: classifier construction and visual classification.

\textbf{Classifier Construction}. 
Similar to CLIP-ReID, we design an initial language description template to represent a pedestrian with a specific identity: ``A $[\text{X}]_1$ $[\text{X}]_2$ ... $[\text{X}]_\text{M}$ person observed in both day and night conditions'', where $[\text{X}]_i$ denotes learnable tokens and $\text{M}$ represents the number of such tokens. Unlike CLIP-ReID, which aligns images and corresponding text within a batch, we process the text prompts for all identities simultaneously during each optimization iteration. Specifically, we input the language descriptions for all identities, $\{T_{i}\}_{i=1}^{N_y}$, into the frozen text encoder $E_t$ to obtain the text features, $\{\textbf{f}_{y_i}^{\hspace{0.1em}t}\}_{i=1}^{N_y}=E_t(\{T_{i}\}_{i=1}^{N_y})$. These text features serve as prototypes for the identity classifier, aiding in the classification of pedestrian sequence-level representations from different modalities.

\textbf{Visual Classification}.
Our goal is to ensure that the $i$-th sequence-level visual feature $\textbf{f}_{m,cls}^{\hspace{0.1em}i,s}$, generated by the visual encoder, is accurately classified by the classifier built using the text prompt learner and the frozen text encoder. To accomplish this goal, we introduce the following visual-to-text loss $L_{v2t}$:
\begin{equation}
\begin{aligned}
\bm L_{v2t} = -\frac{1}{{{n_b}}}\sum\limits_{i = 1}^{{n_b}}  {\log \frac{{\exp \left( {s\left( {\textbf{f}_{m,cls}^{\hspace{0.1em} i,s},\textbf{f}_{t}^{\hspace{0.1em} i}} \right)} \right)}}{{\sum\nolimits_{j = 1}^{{N_y}} {\exp \left( {s\left( {\textbf{f}_{m,cls}^{\hspace{0.1em} i,s},\textbf{f}_{t}^{\hspace{0.1em} j}} \right)} \right)}
}}}   
\end{aligned},
\end{equation}
where $s(\cdot)$ denotes the similarity function between two features and $\exp(\cdot)$ represents the exponential function with the natural base.

Notably, during the entire training process, the weights of the text encoder remain frozen, enabling the training to focus solely on the visual encoder and the prompt learner. This design ensures that, while the visual encoder and prompt learner are continuously updated, the extracted visual features remain consistently aligned with CLIP's original multimodal space.

\subsection{Spatial-Temporal Prompting}\label{sec3.4}
Although IMLP performs well in mitigating modality discrepancies and extracting rich semantic features, it is limited by its reliance on simple temporal average pooling for sequence-level visual feature extraction. This module fails to integrate complementary information across different frames effectively and cannot capture temporal dynamics, such as gait variations. Consequently, the identity-level text prompts generated by IMLP cannot fully represent the rich spatiotemporal information within the sequence, thereby limiting its overall performance. 
To enable interframe interaction, existing methods typically employ LSTMs \cite{li2023intermediary,lin2022learning}, transformers \cite{feng2024cross}, or GCNs \cite{zhou2023video} after the encoder to facilitate information exchange across frames and generate sequence-level features. However, these methods overlook interframe interactions within intermediate network layers and introduce additional networks, resulting in significant computational overhead. 
To address these limitations, we propose \textbf{spatial-temporal prompting (STP)}, which enhances interframe interactions through two key components: \textbf{spatial-temporal hub (STH)} and \textbf{spatial-temporal aggregation (STA)}.

\begin{figure}[t!]
\centering
\includegraphics[width=7cm,keepaspectratio=true]{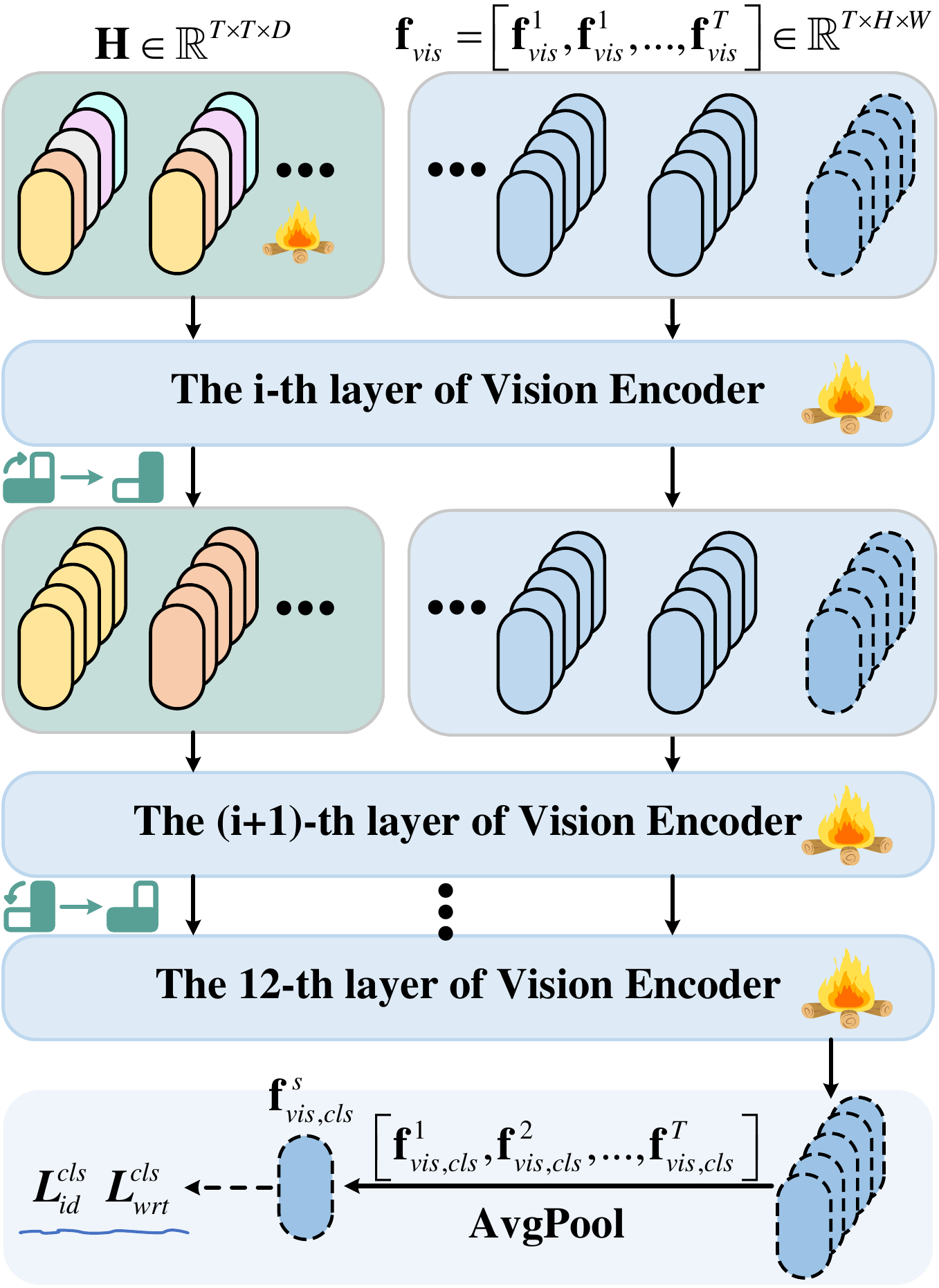}
\caption{Illustration of the proposed STH, which leverages the multihead attention mechanism of the ViT to alternately aggregate and diffuse spatiotemporal information between the \(i\)-th and \(i+1\)-th layers.
}
\label{STH}
\end{figure}

\textbf{Spatial-Temporal Hub}. 
As shown in Fig.\ref{STH}, the spatial-temporal hub (STH) serves as a ``hub" to aggregate spatiotemporal information from different frames and diffuse it back to each frame.
Specifically, taking the visible sequence as an example, given an input sequence \( V \in \mathbb{R}^{T \times H \times W} \), after being divided into \( N \) patches and enhanced with positional encoding, the sequence is represented as \( \mathbf{f}_{vis} = [\mathbf{f}_{vis}^{1}, \mathbf{f}_{vis}^{2}, \dots, \mathbf{f}_{vis}^{T}] \in \mathbb{R}^{T \times (N+1) \times D} \), where \( D \) is the channel dimension, $\textbf{f}_{vis}^{\hspace{0.1em} t} =[\textbf{f}_{vis,cls}^{\hspace{0.1em} t}, \textbf{C}\textbf{f}_{vis}^{\hspace{0.1em} t,1}, \dots, \textbf{C}\textbf{f}_{vis}^{\hspace{0.1em} t,N}] + \text{e}^{spa}$. The STH is then defined as a 3D tensor \( \mathbf{H} \in \mathbb{R}^{T \times T \times D} \), designed to effectively aggregate and diffuse spatiotemporal information. 
Before inputting $\mathbf{f}_{vis}$ into the ViT $E_v$, we concatenate $\mathbf{f}_{vis}$ and $\mathbf{H}$ along the spatial dimension to obtain $VH = [\mathbf{f}_{vis}; \mathbf{H}] \in \mathbb{R}^{T \times (T+N+1) \times D}$. This combined representation is then passed through the $i$-th layer of the $E_v$ that is the starting insertion layer of the STH, which includes a multihead attention (MHA) mechanism $MHA^{(i)}$ followed by a multilayer perceptron (MLP) $MLP^{(i)}$:
% Formally, this process can be expressed as:
\begin{equation}
\begin{aligned}
VH^{\left(i \right)} = MLP^{\left(i \right)}\left(MHA^{\left(i \right)}\left(VH\right)\right)
\end{aligned},
\end{equation}
where $VH^{\left(i \right)}$ represents the result of encoding $VH$ through the $i$-th layer of $E_v$. Owing to the MHA mechanism of the ViT, relationships are established between all patches. Therefore, by concatenating $VH = [\mathbf{f}_{vis}; \mathbf{H}]$, we create a communication channel between $\mathbf{f}_{vis}$ and $\mathbf{H}$, enabling $\mathbf{H}$ to aggregate both global and local spatial information for each frame.

Before inputting \(VH^{(i)} \) into the $(i+1)$-th layer of $E_v$, we perform a transpose operation on \( \mathbf{H} \) in \( VH^{(i)} \) along the spatial and temporal dimensions:
\begin{equation}
\begin{aligned}
VH^{\left(i \right)} = [\mathbf{f}_v; \mathbf{H}] \rightarrow [\mathbf{f}_v; \mathbf{H}^{\top}]
\end{aligned},
\end{equation}
Similar to the process in the $i$-th layer of $E_v$, when \(VH^{(i)} \) is input into the $(i+1)$-th layer of $E_v$, \( VH^{(i+1)} \) can be obtained:
\begin{equation}
\begin{aligned}
VH^{\left(i+1 \right)} = MLP^{\left(i+1 \right)}\left(MHA^{\left(i+1 \right)}\left(VH^{\left(i \right)}\right)\right)
\end{aligned},
\end{equation}
Owing to the transposition operation on $\textbf{H}$, each row of \( \mathbf{H}^{\top} \) of  \( VH^{(i)} \) contains spatiotemporal information, with spatial information from all frames. Through MHA, relationships are established between the \( j \)-th row of \( \mathbf{H}^{\top} \) and the \( j \)-th frame, allowing for the interaction of temporal and spatial information. As a result, \( \mathbf{H} \) acts as a central hub, integrating both spatial and temporal information and diffusing it to each frame. Similarly, before feeding \( VH^{(i)} \) into the subsequent layers of \( E_v \), we alternate the transposition of \( \mathbf{H} \) to facilitate continuous spatiotemporal information exchange across frames.
Ultimately, the spatiotemporal information is transferred to the [CLS] token of each frame, and the sequence-level pedestrian feature $\mathbf{f}_{vis,cls}^{\hspace{0.1em} s}$ is obtained through $\text{AvgPool}$:
\begin{equation}
\begin{aligned}
\mathbf{f}_{vis,cls}^{\hspace{0.1em} s} = \text{AvgPool}([\mathbf{f}_{vis,cls}^{\hspace{0.1em} 1}, \mathbf{f}_{vis,cls}^{\hspace{0.1em} 2}, \dots, \mathbf{f}_{vis,cls}^{\hspace{0.1em} T}]),
\end{aligned},
\end{equation}
where \( [\mathbf{f}_{vis,cls}^{\hspace{0.1em} 1}, \mathbf{f}_{vis,cls}^{\hspace{0.1em} 2}, \dots, \mathbf{f}_{vis,cls}^{\hspace{0.1em} T}] \) represents the [CLS] tokens from \( VH^{(12)} = [\mathbf{f}_{vis}; H] \), the output of the final layer of \( E_v \).

\begin{figure}[t!]
\centering
\includegraphics[width=7cm,keepaspectratio=true]{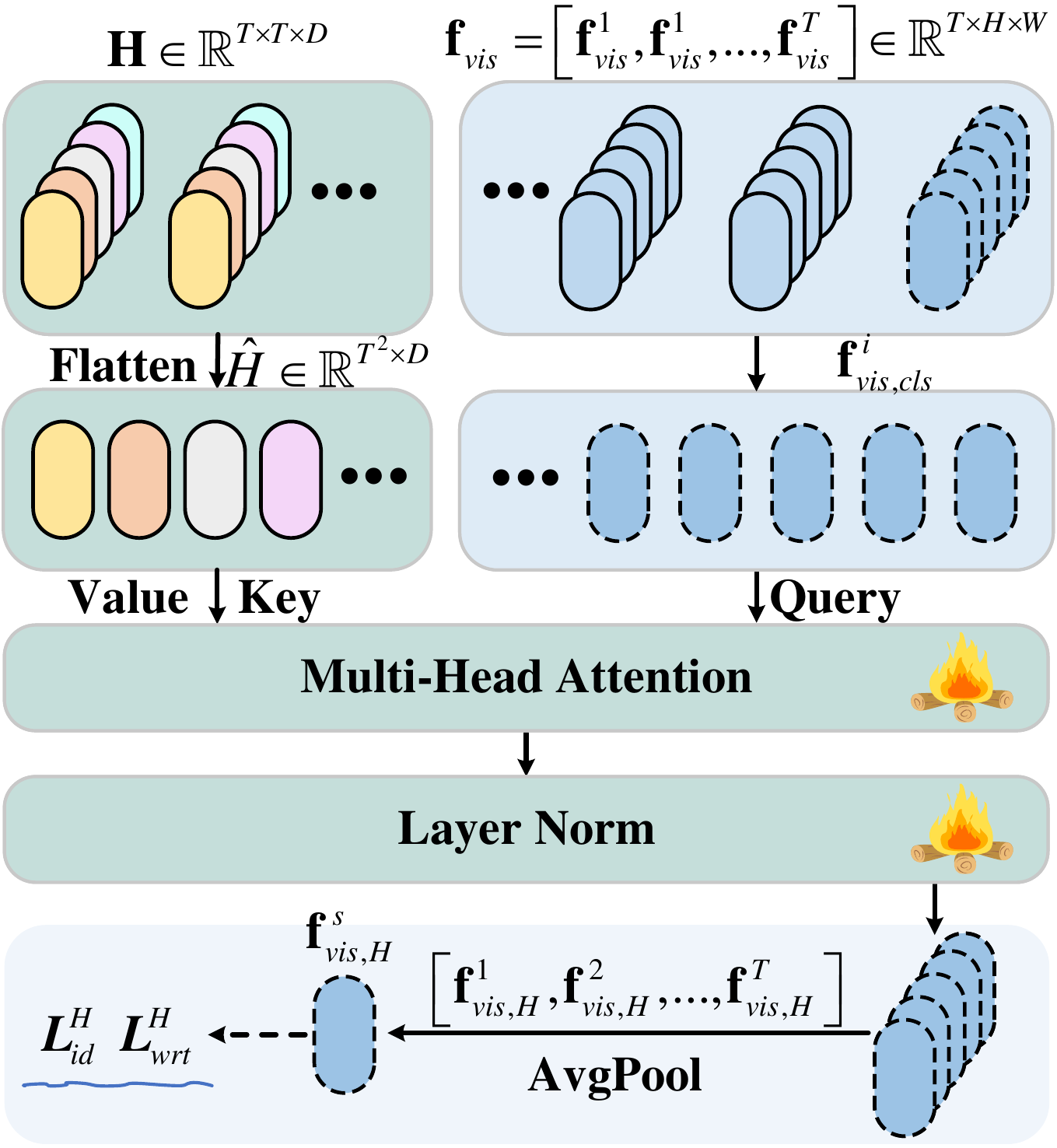}
\caption{An illustration of the proposed STA, which employs specialized multihead attention to establish relationships between the [CLS] token and \(\mathbf{H}\), enhancing the spatiotemporal aggregation capability of the STH.
}
\label{STA}
\end{figure}

\textbf{Spatial-Temporal Aggregation}. 
In the STH, $\mathbf{H}$ acts as a central hub, aggregating spatiotemporal information from multiple frames and diffusing this information to the [CLS] token. However, the aggregation and diffusion processes have certain limitations: first, aggregation cannot ensure that the gathered spatiotemporal information is identity-relevant and free from noise due to the lack of direct loss constraints; second, diffusion is incomplete because the MHA weights in CLIP are not specifically designed to establish a strong connection between the [CLS] token $\mathbf{f}_{vis,cls}^{\hspace{0.1em} i}$ and $\mathbf{H}$, resulting in incomplete transmission of spatiotemporal information. To address these challenges, as shown in Fig.\ref{STA}, we propose \textbf{spatial-temporal aggregation} (\textbf{STA}), which uses specialized multihead attention $MHA_{sta}$ to establish a direct connection between the [CLS] token $\mathbf{f}_{vis,cls}^{\hspace{0.1em} i}$ and $\mathbf{H}$, thus capturing the most identity-relevant spatiotemporal information within $\mathbf{H}$:
\begin{equation}
	\begin{aligned}
	\mathbf{f}_{vis,H}^{\hspace{0.1em} i} =  LN(MHA_{sta}(\mathbf{f}_{vis,cls}^{\hspace{0.1em} i},\hat{\mathbf{H}},\hat{\mathbf{H}}))
	\end{aligned},
\end{equation}
where \( \hat{\mathbf{H}} \in \mathbb{R}^{T^{2} \times D} \) represents the result of flattening \( \mathbf{H} \) along both the spatial and temporal dimensions. In the MHA, \( \hat{\mathbf{H}} \) is used as both the key and value, whereas \( \mathbf{f}_{vis,cls}^{\hspace{0.1em} i} \) serves as the query. $LN$ represents the layer normalization layer.

Furthermore, we apply temporal average pooling $\text{AvgPool}$ to all frame representations $\{\textbf{\text{f}}_{vis,H}^{\hspace{0.1em} i}\}_{i=1}^{T}$ to obtain the sequence-level feature representation:
\begin{equation}
\begin{aligned}
\textbf{f}_{vis,H}^{\hspace{0.1em} s} =\text{AvgPool}([\textbf{f}_{vis,H}^{\hspace{0.1em} 1},\textbf{f}_{vis,H}^{\hspace{0.1em} 2},\dots,\textbf{f}_{vis,H}^{\hspace{0.1em} T}])
\end{aligned},
\end{equation}
To ensure that the spatiotemporal information captured by \( \textbf{H} \) is relevant to pedestrian identity and facilitates better information communication across frames, we introduce cross-entropy loss $\bm L_{id}^{H}$ and weighted regularized triplet loss $\bm L_{wrt}^{H}$ to constrain \( \mathbf{f}_{vis,H}^{\hspace{0.1em} s} \):
\begin{equation}
\begin{aligned}
\bm L_{id}^{H}&=-\frac{1}{n_{b}}\sum_{i=1}^{n_{b}}q_{i}\log(\bm W_{id}^{H}(\textbf{\text{f}}_{m,H}^{\hspace{0.1em} i,s}))
\end{aligned},
\end{equation}
\begin{equation}
\begin{aligned}
% \begin{split}
L_{wrt}^{H}= \frac{1}{n_{b}}\sum_{i=1}^{n_{b}}\log(1+\exp(\sum_{i,j} w_{i,j}^{p}\bm d_{i,j}^{p}-\sum_{i,k} w_{i,k}^{n}\bm d_{i,k}^{n}),\\
w_{i,j}^{p} = \frac{\exp(\bm d_{i,j}^{p})}{\sum_{\bm d_{i,j}\in{P_{i}}}\exp(\bm d_{i,j}^{p})},w_{i,k}^{n} = \frac{\exp(-\bm d_{i,k}^{n})}{\sum_{\bm d_{i,k}\in{N_{i}}}\exp(-\bm d_{i,k}^{n})}
% \end{split}
\end{aligned},
\end{equation}
where \(\textbf{\text{f}}_{m,H}^{\hspace{0.1em} i, s}\) denotes the \(i\)-th sequence-level feature within a batch and $m$ denotes the modality (visible $vis$ or infrared $ir$). $j,k$ ($P_{i},N_{i}$) represents the index of the positive and negative samples (set) corresponding to the anchor sample within a batch size. $\bm{d}_{i,j}$ represents the Euclidean distance between two features: $\bm d_{i,j}  =\bm \|\textbf{\text{f}}_{m,H}^{\hspace{0.1em} i,s}-\textbf{\text{f}}_{m,H}^{\hspace{0.1em} j,s}\|_{2}$.

% \textbf{Discussion:}
\subsection{Discussion}\label{sec3.4.1}
We categorize our contributions into two complementary aspects: combination novelty, which unifies diverse paradigms in a novel architecture, and incremental novelty, which introduces two new modules that solve key challenges in VVI-ReID.

\textbf{Combination Novelty.} Rather than simply stacking existing components, VLD is the first framework to systematically integrate CLIP, language prompting, cross-modal alignment, and video-level spatiotemporal modeling into a unified, end-to-end trainable architecture. This joint design enables modality-invariant and identity-discriminative representations to be learned in a single pipeline. Notably, to the best of our knowledge, this is the first work to extend CLIP to the VVI-ReID task, where multimodal and multiframe challenges coexist.

\textbf{Incremental Novelty.} VLD introduces two novel components that contribute to distinct technical advances. First, the invariant-modality language-prompting (IMLP) module treats learnable prompts as identity prototypes and formulates a visual-to-text alignment loss to bridge the modality gap within CLIP's embedding space. This step effectively transforms the frozen text encoder and identity-level prompts into a modality-invariant classifier. Second, the spatial-temporal prompting (STP) module enhances spatiotemporal modeling by injecting a learnable spatiotemporal hub (STH) across intermediate layers of the vision transformer (ViT). Together with spatial-temporal aggregation (STA), STP enables rich interframe interactions and identity-specific attention without introducing significant computational overhead, requiring only +2.39M parameters and +0.12G FLOPs.

\subsection{Optimization}\label{sec3.5}
The training process is end-to-end. We introduce \(\bm L_{id}^{cls}\) and \(\bm L_{wrt}^{cls}\) to ensure that sequence-level pedestrian representations maintain identity discriminability. Similarly, \(\bm L_{id}^{H}\) and \(\bm L_{wrt}^{H}\) ensure that the spatiotemporal information aggregated by STH remains relevant to identity. Additionally, \(\bm L_{v2t}\) learns a modality-shared identity-level text prompt to describe both visible and infrared pedestrian sequences, reducing modality gaps and enhancing semantic representation. The total loss is defined as:
\begin{equation}
	\begin{aligned}
	\bm L_{total} = \bm L_{id}^{cls}  + \bm L_{wrt}^{cls} +  \lambda_1 \bm L_{v2t}+   \lambda_2 \bm L_{id}^{H}  +  \lambda_3 \bm L_{wrt}^{H} 
	\end{aligned},
\end{equation}
where $\lambda_1$, $\lambda_2$ and $\lambda_3$ are hyperparameters used to balance the loss terms.

\begin{table*}[!ht]\centering\small
\caption{Values of mAP and CMC (\%) obtained by our proposed method and the state-of-the-art Re-ID methods on HITSZ-VCM. ``R@1'', ``R@5'' and ``R@10'' denote Rank-1, Rank-5 and Rank-10, respectively. All results are taken from the corresponding papers. †: Results obtained using the official implementation.}
\label{vcm}
\begin{tabular}{m{2.3cm}<{\centering}m{2cm}<{\centering}m{1.5cm}<{\centering}m{1cm}<{\centering}m{0.7cm}<{\centering}m{0.7cm}<{\centering}m{0.7cm}<{\centering}m{0.7cm}<{\centering}m{0.7cm}<{\centering}m{0.7cm}<{\centering}m{0.7cm}<{\centering}m{0.7cm}<{\centering}}
\toprule[0.8pt]
% \rowcolor[HTML]{f5e9fb}
\begin{tikzpicture}[overlay, remember picture]
    \fill[gray!30] (-0.7,-0.6) rectangle (16.76,0.31); % adjust coordinates to match your table cells
\end{tikzpicture}
\begin{tikzpicture}[overlay, remember picture]
    \fill[gray!10] (-0.82,-5.41) rectangle (16.65,-6.05); % adjust coordinates to match your table cells
\end{tikzpicture}
\multirow{2}*{Methods} & \multirow{2}*{Reference} & \multirow{2}*{Type} & \multirow{2}*{Seq\_Len} & \multicolumn{4}{c}{\textit{Infrared-to-Visible}} & \multicolumn{4}{c}{\textit{Visible-to-Infrared}}  
\\ \cmidrule(lr){5-8} \cmidrule(lr){9-12} 
% \rowcolor[HTML]{f5e9fb}
                            &  &   &  & {R@1}  & {R@5}  & {R@10}  & mAP    & {R@1}  & {R@5} & {R@10}  & mAP           \\ \toprule[0.8pt]

% \multirow{5}*{Image} 
 Lba\cite{park2021learning}       &ICCV'21 &Image  &6    &46.4 &65.3 &72.2 &30.7 &49.3 &69.3 &75.9 &32.4  \\
 MPANet\cite{wu2021discover}       &CVPR'21 &Image  &6    &46.5 &63.1 &70.5 &35.3 &50.3 &67.3 &73.6 &37.8 \\
 VSD\cite{tian2021farewell}        &CVPR'21 &Image  &6    &54.5 &70.0 &76.3 &41.2 &57.5 &73.7 &79.4 &43.5  \\
 CAJ\cite{ye2021channel}           &ICCV'21 &Image  &6    &56.6 &73.5 &79.5 &41.5 &60.1 &74.6 &79.9 &42.8  \\
 SEFL\cite{feng2023shape}          &CVPR'23 &Image  &6    &67.7 &80.3 &84.7 &52.3 &70.2 &82.2 &86.1 &52.5 \\
  CLIP-ReID† \cite{li2023clip}       &AAAI'23 &Image  &6    &58.4 &73.2 &79.8 &45.3 &60.4 &76.9 &83.7 &43.5 \\
    TF-CLIP† \cite{yu2023tf}       &AAAI'24 &Image  &6    &62.3 &76.2 &81.6 &47.5 &62.2 &79.6 &85.5 &45.5 \\
 \hline
 % \multirow{6}*{Video} 
 MITML\cite{lin2022learning}      &CVPR'22  &Video &6    &63.7 &76.9 &81.7  &45.3 &64.5 &79.0 &83.0 &47.7   \\
 IBAN\cite{li2023intermediary}    &TCSVT'23 &Video &6    &65.0 &78.3 &83.0  &48.8 &69.6 &81.5 &85.4 &51.0   \\
 SADSTRM\cite{li2023adversarial}  &Arxiv'23 &Video &6    &65.3 &77.9 &82.7  &49.5 &67.7 &80.7 &85.1 &51.8   \\
 SAADG\cite{zhou2023video}        &ACM MM'23 &Video &6    &\uwave{69.2} &\uwave{80.6} &\uwave{85.0}  &\uwave{53.8} &\uwave{73.1} &\uwave{83.5} &\uwave{86.9} &\uwave{56.1}   \\
 CST\cite{feng2024cross}          &TMM'24  &Video  &6    &69.4 &81.1 &85.8  &51.2 &72.6 &83.4 &86.7 &53.0   \\
 AuxNet\cite{du2023video}         &TIFS'24 &Video  &6     &51.1 &- &-  &46.0 &54.6 &- &- &48.7   \\
 \hline
 
 \textbf{our}     &-                       &Video   &6    & \textbf{74.3} & \textbf{85.0} & \textbf{88.4} & \textbf{60.2}  & \textbf{74.6} & \textbf{86.4} & \textbf{90.0} & \textbf{58.6} \\
 \textbf{our}     &-                       &Video   &10   &\textbf{74.2} &\textbf{85.8} &\textbf{89.3} &\textbf{61.8}     &\textbf{75.6} &\textbf{86.4} &\textbf{90.2} &\textbf{60.0}   \\ \toprule[0.8pt]
\end{tabular}
\end{table*}
\begin{table*}[!ht]\centering\small
\caption{Values of mAP and CMC (\%) obtained by our proposed method and the state-of-the-art Re-ID methods on BUPTCampus. ``R@1'', ``R@5'' and ``R@10'' denote Rank-1, Rank-5 and Rank-10, respectively. All results are taken from the corresponding papers. †: Results obtained using the official implementation.
}
\label{bupt}
\begin{tabular}{m{2.3cm}<{\centering}m{2cm}<{\centering}m{1.5cm}<{\centering}m{1cm}<{\centering}m{0.7cm}<{\centering}m{0.7cm}<{\centering}m{0.7cm}<{\centering}m{0.7cm}<{\centering}m{0.7cm}<{\centering}m{0.7cm}<{\centering}m{0.7cm}<{\centering}m{0.7cm}<{\centering}}
\toprule[0.8pt]
\begin{tikzpicture}[overlay, remember picture]
    \fill[gray!30] (-0.7,-0.6) rectangle (16.77,0.31); % adjust coordinates to match your table cells
\end{tikzpicture}
\begin{tikzpicture}[overlay, remember picture]
    \fill[gray!10] (-0.82,-5.06) rectangle (16.65,-5.69); % adjust coordinates to match your table cells
\end{tikzpicture}
\multirow{2}*{Methods} & \multirow{2}*{Reference} & \multirow{2}*{Type} & \multirow{2}*{Seq\_Len} & \multicolumn{4}{c}{\textit{Infrared-to-Visible}} & \multicolumn{4}{c}{\textit{Visible-to-Infrared}} \\ 
\cmidrule(lr){5-8} \cmidrule(lr){9-12} 
% \rowcolor{gray!30}
& & & & {R@1} & {R@5} & {R@10} & mAP & {R@1} & {R@5} & {R@10} & mAP \\ \toprule[0.8pt]

% \multirow{8}*{Image} 
  AlignGAN\cite{wang2019rgb}    &ICCV'19 &Image  &10  & 28.0 & 49.1 & 57.7  & 30.3 & 35.4 & 53.9 & 61.3 & 35.1  \\
    DDAG\cite{ye2020dynamic}      &ECCV'20 &Image  &10  & 40.4 & 61.4 & 69.8  & 40.4 & 46.3 & 68.2 & 74.4  & 43.1  \\
  LbA\cite{park2021learning}    &ICCV'21  &Image &10  & 32.1 & 54.9 & 65.1  & 32.9 & 39.1 & 58.7 & 66.5  & 37.1 \\

  CAJ\cite{ye2021channel}       &ICCV'21 &Image  &10  & 40.5 & 66.8 & 73.3  & 41.5 & 45.0 & 70.0 & 77.0  & 43.6  \\
  AGW\cite{ye2021deep}          &TPAMI'21 &Image &10  & 36.4 & 60.1 & 67.2  & 37.4 & 43.7 & 64.4 & 73.2  & 41.1 \\
  MMN\cite{zhang2021towards}    &CVPR'21 &Image  &10  & 40.9 & 67.2 & 74.4  & 41.7 & 43.7 & 65.2 & 73.5  & 42.8 \\
    DART\cite{yang2022learning}   &CVPR'22 &Image  &10  &52.4 &70.5 &77.8 &49.1  &53.3  &75.2  &81.7  &50.5\\
  DEEN\cite{zhang2023diverse}   &CVPR'23  &Image &10  &53.7 &74.8 &80.7 &50.4  &49.8  &71.6  &81.0  &48.6 \\

CLIP-ReID†\cite{li2023clip}   &AAAI'23 &Image  &6  &49.0 &73.0 &81.2 &50.4  &51.0  &75.4  &80.0  &49.8\\
TF-CLIP†\cite{yu2023tf}   &AAAI'24 &Image  &6    &49.4  &76.8  &83.7  &51.9 &52.5 &75.2 &81.5 &51.8\\
 \hline
 % \multirow{2}*{Video} 
 MITML\cite{lin2022learning}    &CVPR'22  &Video &6  &49.1 &67.9 &75.4 &47.5 &50.2 &68.3 &75.7 &46.3   \\
 AuxNet\cite{du2023video}       &TIFS'24  &Video &10  &\uwave{63.6} &\uwave{79.9} &\uwave{85.3} &\uwave{61.1} &\uwave{62.7} &\uwave{81.5} &\uwave{85.7} &\uwave{60.2}    \\
 \hline

    % \textbf{our}     &-               &Video  &6   &\textbf{66.7} &\textbf{85.2} &\textbf{91.0} &\textbf{64.7} &\textbf{65.6} &\textbf{83.8} &\textbf{88.5} &\textbf{62.7}   \\
    \textbf{our}     &-               &Video  &6   &\textbf{65.3} &\textbf{84.9} &\textbf{89.7} &\textbf{63.5} &\textbf{65.8} &\textbf{83.0} &\textbf{87.9} &\textbf{63.0}   \\
   \textbf{our}     &-               &Video  &10   &\textbf{66.7} &\textbf{85.8} &\textbf{89.9} &\textbf{66.0}     &\textbf{67.4} &\textbf{84.2} &\textbf{89.1} &\textbf{64.1}   \\ \toprule[0.8pt]
\end{tabular}
\end{table*}

\section{Experiments}
\subsection{Datasets and Evaluation Metrics}
\noindent\textbf{HITSZ-VCM} \cite{lin2022learning} contains images captured by 6 visible and 6 infrared cameras, including 927 identities, 251,452 RGB images, and 211,807 infrared images. Each tracklet comprises 24 consecutive frames. The dataset is split into a training set with 500 identities and a testing set with 427 identities.

\noindent\textbf{BUPTCampus} \cite{du2023video} includes images captured by 6 cameras, featuring 3,080 identities and 16,826 video tracklets with 1,869,366 images. It is divided into three subsets: 1,074 identities for primary learning, 930 for auxiliary learning, and 1,076 for testing.

\noindent\textbf{Evaluation Metrics}.  
Retrieval performance is measured using cumulative matching characteristic (CMC) and the mean average precision (mAP). Following \cite{lin2022learning,du2023video}, two evaluation strategies are used: infrared-to-visible, where infrared tracklets are the query set and visible tracklets are the gallery set, and visible-to-infrared, where the roles are reversed.

\vspace{-6.5pt}
\subsection{Implementation Details}
The proposed VLD framework was implemented using the PyTorch deep learning platform and tested on a single A6000 GPU. 
To eliminate the influence of randomness, all the experiments were conducted with a fixed random seed (set as 1), thus ensuring reproducibility across runs. Specifically, we fixed the random seed for Python, NumPy, and PyTorch and maintained consistent settings for dataset splits, network initialization, and the data loading order.
The model uses the CLIP vision encoder (ViT-B-16) and text encoder to extract visual and textual features. The input images were resized to \(288 \times 144\) with standard data augmentation techniques, including random horizontal flipping, padding, cropping, channel erasure, and channel swapping \cite{ye2021channel}. 
The Adam optimizer was used with a base learning rate of \(2.5 \times 10^{-5}\), while the learning rate for the prompt learners was set to 25 times the base rate. A cosine learning rate decay strategy \cite{he2021transreid} was used to dynamically adjust the rates. The training spanned 24 epochs with a batch size of 32, comprising 2 modalities with 4 pedestrians each. Each pedestrian had 4 sequences, and 6 images were randomly sampled per sequence. The number of learnable tokens $\text{M}$ was set to 4, with hyperparameters $\lambda_1$, $\lambda_2$ and $\lambda_3$ set to 0.08, 0.4 and 1, respectively.

% \vspace{-4pt}
\subsection{Comparison with State-of-the-Art Methods}
In this section, we conducted comparative experiments with other state-of-the-art methods on the HITSZ-VCM and BUPTCampus datasets. Specifically, the comparison methods include the following short-term person ReID methods: 
Lba\cite{park2021learning}, 
MPANet\cite{wu2021discover},  
VSD\cite{tian2021farewell},  
CAJ\cite{ye2021channel},   
SEFL\cite{feng2023shape},  
MITML\cite{lin2022learning},     
IBAN\cite{li2023intermediary},
SADSTRM\cite{li2023adversarial}, 
SAADG\cite{zhou2023video}, 
CST\cite{feng2024cross}, 
AlignGAN\cite{wang2019rgb}, DDAG\cite{ye2020dynamic}, 
 AGW\cite{ye2021deep}, MMN\cite{zhang2021towards}, DEEN\cite{zhang2023diverse}, DART\cite{yang2022learning}, MITML\cite{lin2022learning}, AuxNet\cite{du2023video},

\begin{table*}[!ht]\small
 \centering {\caption{Ablation studies of the proposed VLD. ``B'': Baseline. ``IMLP'': Invariant-
Modality Text Prompting. ``STP'': Spatial-Temporal Prompting. }\label{component}
\begin{tikzpicture}[overlay, remember picture]
    \fill[gray!30] (0.22,0.44) rectangle (14.08,1.35); % adjust coordinates to match your table cells
\end{tikzpicture}
\begin{tikzpicture}[overlay, remember picture]
    \fill[gray!10] (0.11,-0.77) rectangle (13.95,-1.1); % adjust coordinates to match your table cells
\end{tikzpicture}
\begin{tabular}{m{1.1cm}<{\centering}m{1.3cm}<{\centering}m{1.3cm}<{\centering}m{1.8cm}<{\centering}m{1.8cm}<{\centering}m{1.8cm}<{\centering}m{1.8cm}<{\centering}}
\toprule[0.8pt] 
\multicolumn{3}{c}{Component}  & \multicolumn{2}{c}{\textit{Infrared-to-Visible}}   & \multicolumn{2}{c}{\textit{Visible-to-Infrared}} \\ 
% \cline{2-5} 
\cmidrule(lr){1-3} \cmidrule(lr){4-5} \cmidrule(lr){6-7} 
% \cline{6-7} \cline{8-9}
  B &IMLP & STP      & R@1    & mAP & R@1    & mAP \\ 
\toprule[0.8pt]
\textcolor[rgb]{0,0.5,0}{\ding{51}} &\textcolor{red}{\ding{55}}   &\textcolor{red}{\ding{55}}          & 67.0  & 52.6  &64.2  &49.3\\
 \textcolor[rgb]{0,0.5,0}{\ding{51}} &\textcolor[rgb]{0,0.5,0}{\ding{51}} &\textcolor{red}{\ding{55}}         & 68.1\textcolor[rgb]{0,0.5,0}{(+1.1)}  & 54.6\textcolor[rgb]{0,0.5,0}{(+2.0)}  & 67.2\textcolor[rgb]{0,0.5,0}{(+3.0)}  & 51.0\textcolor[rgb]{0,0.5,0}{(+1.7)} \\
 \textcolor[rgb]{0,0.5,0}{\ding{51}} &\textcolor{red}{\ding{55}}  &\textcolor[rgb]{0,0.5,0}{\ding{51}}       & 72.1\textcolor[rgb]{0,0.5,0}{(+5.1)}  & 58.2\textcolor[rgb]{0,0.5,0}{(+5.6)}  & 73.2\textcolor[rgb]{0,0.5,0}{(+9.0)}  & 57.2\textcolor[rgb]{0,0.5,0}{(+7.9)} \\
 \textcolor[rgb]{0,0.5,0}{\ding{51}} &\textcolor[rgb]{0,0.5,0}{\ding{51}} &\textcolor[rgb]{0,0.5,0}{\ding{51}}      & 74.3\textcolor[rgb]{0,0.5,0}{(+7.3)}  & 60.2\textcolor[rgb]{0,0.5,0}{(+7.6)}  & 74.6\textcolor[rgb]{0,0.5,0}{(+10.4)}  & 58.6\textcolor[rgb]{0,0.5,0}{(+9.3)} \\
\toprule[0.8pt]
\end{tabular}}
\end{table*}

\begin{table*}[!ht]\small
 \centering {\caption{Comparison of different temporal
fusion methods, where IR represents infrared and VIS represents visible.}\label{sfm}
% \begin{tikzpicture}[overlay, remember picture]
%     \fill[gray!30] (0.1,0.27) rectangle (12.5,1.18); % adjust coordinates to match your table cells
% \end{tikzpicture}  &FLOPs(G
\begin{tikzpicture}[overlay, remember picture]
    \fill[gray!30] (0.22,0.63) rectangle (13.99,1.53); % adjust coordinates to match your table cells
\end{tikzpicture}
\begin{tikzpicture}[overlay, remember picture]
    \fill[gray!10] (0.11,-0.95) rectangle (13.87,-1.28); % adjust coordinates to match your table cells
\end{tikzpicture}
\begin{tabular}{m{1.2cm}<{\centering}m{1.4cm}<{\centering}m{1.4cm}<{\centering}m{1.4cm}<{\centering}m{1.4cm}<{\centering}m{2cm}<{\centering}m{2cm}<{\centering}}
\toprule[0.8pt]
  \multirow{2}*{Methods}  & \multicolumn{2}{c}{\textit{IR-to-Vis}}   & \multicolumn{2}{c}{\textit{Vis-to-IR}}  &\multirow{2}*{Params/M} &\multirow{2}*{FLOPs/G}\\ 
% \cmidrule(lr){1} 
\cmidrule(lr){2-3} \cmidrule(lr){4-5} 
      & R@1    & mAP & R@1  & mAP &  & \\ 
\toprule[0.8pt]
 TAP            & 68.1  & 54.6  & 67.2  & 51.0 &86.17 &13.96\\
Conv1D          & 66.3\textcolor[rgb]{1,0,0}{(-1.8)}  & 53.7\textcolor[rgb]{1,0,0}{(-0.9)}  & 67.1\textcolor[rgb]{1,0,0}{(-0.1)}  & 51.4\textcolor[rgb]{1,0,0}{(+0.4)} &86.21\textcolor[rgb]{1,0,0}{(+0.04)} &13.96\textcolor[rgb]{1,0,0}{(+0.00)}\\
Transf          & 64.5\textcolor[rgb]{1,0,0}{(-3.6)}  & 50.9\textcolor[rgb]{1,0,0}{(-3.7)}  & 61.8\textcolor[rgb]{1,0,0}{(-5.4)}  & 47.5\textcolor[rgb]{1,0,0}{(-3.5)} &105.08\textcolor[rgb]{1,0,0}{(+18.91)} &14.00\textcolor[rgb]{1,0,0}{(+0.04)}\\
$\text{Transf}_{cls}$          
               & 69.9\textcolor[rgb]{0,0.5,0}{(+1.8)}  & 56.3\textcolor[rgb]{0,0.5,0}{(+1.7)}  & 71.1\textcolor[rgb]{0,0.5,0}{(+3.9)}  & 54.9\textcolor[rgb]{0,0.5,0}{(+3.9)} &105.13\textcolor[rgb]{0,0.5,0}{(+18.96)} &14.00\textcolor[rgb]{0,0.5,0}{(+0.04)}\\
% TMD            & 62.3  & 47.5  & 62.2  & 45.5 &104.26 &15.16\\
STP            & 74.3\textcolor[rgb]{0,0.5,0}{(+6.2)}  & 60.2\textcolor[rgb]{0,0.5,0}{(+5.6)}   & 74.6\textcolor[rgb]{0,0.5,0}{(+7.4)}  & 58.6\textcolor[rgb]{0,0.5,0}{(+7.6)} &88.56\textcolor[rgb]{0,0.5,0}{(+2.39)} &14.08\textcolor[rgb]{0,0.5,0}{(+0.12)}\\

% % \hline
% MITML    & 63.7  & 45.3  & 64.5  & 47.7 &159.12/12.20\\
% VLD      & 74.3  & 60.2  & 74.6  & 58.6 &88.56/14.08\\

\toprule[0.8pt]
\end{tabular}}
\end{table*}

\textbf{Evaluation on the HITSZ-VCM Dataset.}
The proposed method was evaluated on the HITSZ-VCM benchmark, which demonstrated significant performance improvements over state-of-the-art methods in both Rank-1 accuracy and mAP, as shown in Tab. \ref{vcm}.
For the infrared-to-visible evaluation with a sequence length of 6, our method achieved a Rank-1 accuracy of 74.3\%, surpassing the previous best, SAADG (69.2\%). It also achieved a mAP of 60.2\%, outperforming that of SAADG by 53.8\%. In the visible-to-infrared evaluation, our method reached a Rank-1 accuracy of 74.6\% and a mAP of 58.6\%, exceeding SAADG's corresponding values of 73.1\% and 56.1\%, respectively.
With a sequence length of 10 (Tab. \ref{vcm}), the infrared-to-visible evaluation achieved a Rank-1 accuracy of 74.2\% and a mAP of 61.8\%. For the visible-to-infrared evaluation, the Rank-1 accuracy increased to 75.6\%, with a mAP of 60.0\%.
These results highlight the superior performance of our method across both evaluation protocols, which consistently surpasses existing methods in both Rank-1 accuracy and mAP.

\textbf{Evaluation on the BUPTCampus Dataset.}
% Evaluation on BUPTCampus Dataset  
To evaluate scalability, we conducted experiments on the BUPTCampus benchmark, a larger and more diverse pedestrian dataset. As shown in Tab. \ref{bupt}, our method significantly outperforms state-of-the-art methods, even with a sequence length of 6.
In the infrared-to-visible evaluation, our method achieved a Rank-1 accuracy of 65.3\% and a mAP of 63.5\%, surpassing AuxNet by 1.7\% and 2.4\%, respectively. The visible-to-infrared evaluation achieved a Rank-1 accuracy of 65.8\% and a mAP of 63.0\%, outperforming AuxNet (using a sequence length of 10) by 3.1\% and 2.8\%, respectively.
With a sequence length of 10, our method is further improved. In the infrared-to-visible evaluation, the Rank-1 accuracy and mAP increased to 66.7\% and 66.0\%, respectively. In the visible-to-infrared evaluation, the Rank-1 accuracy reached 67.4\%, with a mAP of 64.1\%.
These results demonstrate the robustness of our method in capturing modality-invariant sequence-level pedestrian features. By effectively utilizing spatiotemporal information within sequences and aligning features in the CLIP embedding space, it successfully mitigates modality differences in the sequence.

\begin{table*}[!ht]\small
 \centering {\caption{Ablation studies of the proposed STP. ``B'': Baseline. ``IMLP'': Invariant-
Modality Text Prompting. ``STH'': Spatial-Temporal Hub
. ``STA'': Spatial-Temporal Aggregation
.}\label{STH_componenet}
\begin{tikzpicture}[overlay, remember picture]
    \fill[gray!30] (0.2,0.27) rectangle (14.08,1.18); % adjust coordinates to match your table cells
\end{tikzpicture}
\begin{tikzpicture}[overlay, remember picture]
    \fill[gray!10] (0.11,-0.6) rectangle (13.95,-0.93); % adjust coordinates to match your table cells
\end{tikzpicture}
\begin{tabular}{m{1.1cm}<{\centering}m{1.3cm}<{\centering}m{1.3cm}<{\centering}m{1.8cm}<{\centering}m{1.8cm}<{\centering}m{1.8cm}<{\centering}m{1.8cm}<{\centering}}
\toprule[0.8pt]
  \multicolumn{3}{c}{Component}  & \multicolumn{2}{c}{\textit{Infrared-to-Visible}}   & \multicolumn{2}{c}{\textit{Visible-to-Infrared}} \\ 
\cmidrule(lr){1-3} \cmidrule(lr){4-5} \cmidrule(lr){6-7} 
  B+IMLP &STH & STA      & R@1    & mAP & R@1    & mAP \\ 
\toprule[0.8pt]
 \textcolor[rgb]{0,0.5,0}{\ding{51}} &\textcolor{red}{\ding{55}}   &\textcolor{red}{\ding{55}}         & 68.1  & 54.6  & 67.2  & 51.0 \\
 % \textcolor[rgb]{0,0.5,0}{\ding{51}} &\textcolor[rgb]{0,0.5,0}{\ding{51}} &\textcolor{red}{\ding{55}}       & 71.1\textcolor[rgb]{0,0.5,0}{(+3.4)}  & 57.3\textcolor[rgb]{0,0.5,0}{(+2.5)}  & 71.0\textcolor[rgb]{0,0.5,0}{(+3.9)}  & 54.6\textcolor[rgb]{0,0.5,0}{(+2.1)} \\
 \textcolor[rgb]{0,0.5,0}{\ding{51}} &\textcolor[rgb]{0,0.5,0}{\ding{51}} &\textcolor{red}{\ding{55}}       & 69.6\textcolor[rgb]{0,0.5,0}{(+1.5)}  & 56.9\textcolor[rgb]{0,0.5,0}{(+2.3)}  & 70.7\textcolor[rgb]{0,0.5,0}{(+3.5)}  & 54.2\textcolor[rgb]{0,0.5,0}{(+3.2)} \\
\textcolor[rgb]{0,0.5,0}{\ding{51}} &\textcolor[rgb]{0,0.5,0}{\ding{51}} &\textcolor[rgb]{0,0.5,0}{\ding{51}}     & 74.3\textcolor[rgb]{0,0.5,0}{(+6.2)}  & 60.2\textcolor[rgb]{0,0.5,0}{(+5.6)}   & 74.6\textcolor[rgb]{0,0.5,0}{(+7.4)}  & 58.6\textcolor[rgb]{0,0.5,0}{(+7.6)} \\
\toprule[0.8pt]
\end{tabular}}
\end{table*}

\begin{table}[!t]\small
 \centering {\caption{Ablation studies of the insertion layer of the STH}\label{inser_layer}
\begin{tabular}{m{2cm}<{\centering}m{0.8cm}<{\centering}m{0.8cm}<{\centering}m{0.8cm}<{\centering}m{0.8cm}<{\centering}}
\toprule[0.8pt]
\begin{tikzpicture}[overlay, remember picture]
    \fill[gray!30] (-0.25,-0.6) rectangle (7.78,0.31); % adjust coordinates to match your table cells
\end{tikzpicture}
  \multirow{2}*{Insertion layer}  & \multicolumn{2}{c}{\textit{Infrared-to-Visible}}   & \multicolumn{2}{c}{\textit{Visible-to-Infrared}} \\ 
% \cline{2-5} 
\cmidrule(lr){2-3}  \cmidrule(lr){4-5}
% \cline{6-7} \cline{8-9}
      & R@1    & mAP & R@1    & mAP \\ 
\toprule[0.8pt]
0       & 73.8  & 60.2  &73.4    &58.2\\
1       & 73.1  & 59.8  &73.5    &57.9\\
2       & 73.3  & 59.6  &72.4    &57.3\\
3       & 73.9  & 60.2  &73.4    &57.9\\
4       & 73.0  & 59.5  &72.1    &57.2\\
5       & 73.9  & 60.2  &73.4    &57.9\\
6       & 73.7  & 60.0  &72.7    &57.5\\
7       & 73.7  & 60.1  &73.5    &58.0\\
8       & 73.6  & 60.1  &73.5    &57.9\\
\begin{tikzpicture}[overlay, remember picture]
    \fill[gray!10] (-1.1,-0.1) rectangle (6.9,0.31); % adjust coordinates to match your table cells
\end{tikzpicture}
\textbf{9}       & \textbf{74.4}  & \textbf{60.6}  &\textbf{74.6}    &\textbf{58.6}\\
10      & 73.1  & 59.8  &73.5    &57.9\\

\toprule[0.8pt]
\end{tabular}}
\end{table}

% \vspace{-4pt}
\subsection{Ablation Study}
To evaluate the effectiveness of each component in our VLD, we performed ablation experiments on the HITSZ-VCM dataset. The baseline trains the CLIP vision encoder using only \(\bm{L}_{id}^{cls}\) and \(\bm{L}_{wrt}^{cls}\) losses. 

% The evaluated components include Invariant-Modality Language Prompting (IMLP) and Spatial-Temporal Prompting (STP).

\textbf{\textit{1) Analysis of IMLP.}}  
To evaluate the effectiveness of the IMLP module, we integrate it into the baseline. As shown in Tab. \ref{component}, the addition of IMLP led to significant performance improvements. In the infrared-to-visible evaluation, the Rank-1 accuracy increased from 67.0\% to 68.1\% (+1.1\%), and the mAP improved from 52.6\% to 54.6\% (+2.0\%). Similarly, in the visible-to-infrared evaluation, the Rank-1 accuracy increased from 64.2\% to 67.2\% (+3.0\%), and the mAP increased from 49.3\% to 51.0\% (+1.7\%). These results confirm that the IMLP module effectively reduces modality differences and enhances feature consistency across modalities.

Furthermore, IMLP was integrated into ``B+STP,'' which already incorporates spatiotemporal information modeling. As shown in Tab. \ref{component}, in the infrared-to-visible evaluation, the Rank-1 accuracy increased from 72.1\% to 74.3\% (+2.2\%), and the mAP improved from 58.2\% to 60.2\% (+2.0\%). In the visible-to-infrared evaluation, the Rank-1 accuracy increased from 73.2\% to 74.6\% (+1.4\%), and the mAP increased from 57.2\% to 58.6\% (+1.4\%). These results demonstrate that the text prompts generated by IMLP, enriched with spatiotemporal information, effectively mitigate modality differences in video sequences and enhance the model's cross-modality matching capabilities.

\textbf{\textit{2) Analysis of STP.}}
To evaluate the effectiveness of the STP module, we integrated it into the baseline. As shown in Tab. \ref{component}, in the infrared-to-visible evaluation, the Rank-1 accuracy increased from 67.0\% to 72.1\% (+5.1\%), and the mAP improved from 52.6\% to 58.2\% (+5.6\%). Similarly, in the visible-to-infrared evaluation, the Rank-1 accuracy increased from 64.2\% to 73.2\% (+9.0\%), and the mAP increased from 49.3\% to 57.2\% (+7.9\%). These results confirm that the STP module effectively captures spatiotemporal information, enabling comprehensive feature representation.

When combined with ``B+IMLP'' to form ``B+IMLP+STP,'' performance improved further. As shown in Tab. \ref{component}, in the infrared-to-visible evaluation, the Rank-1 accuracy increased from 68.1\% to 74.3\% (+6.2\%), and the mAP improved from 54.6\% to 60.2\% (+5.6\%). These results validate the complementary nature of STP and IMLP, highlighting how IMLP generates text prompts that effectively encode spatiotemporal information, whereas the STP module plays a crucial role in enhancing spatiotemporal modeling and cross-modality matching performance.

\textbf{Comparison of different temporal aggregation methods.} 
To assess the impact of different temporal aggregation strategies, we replaced TAP with Conv1D, Transf, $\text{Transf}_{cls}$, and STP, drawing inspiration from TF-CLIP\cite{yu2023tf}. As shown in Tab. \ref{sfm}, Conv1D and Transf reduce performance; Conv1D causes a slight drop and minimal resource impact, while Transf significantly degrades performance and increases parameters and computation, making it unsuitable for this task.
In contrast, $\text{Transf}_{cls}$ and STP enhance performance. $\text{Transf}_{cls}$ achieves significant improvements in the infrared-to-visible and visible-to-infrared evaluation but incurs high resource costs. STP, however, provides the best performance boost with minimal overhead. For example, in the infrared-to-visible evaluatio, Rank-1 improves by 6.2\%, and mAP, by 5.6\%, with only 2.39M additional parameters and 0.12G more computations.
STP's effectiveness stems from STH leveraging ViT's attention mechanism in intermediate layers to aggregate spatiotemporal information, further enhanced by STA, thus achieving an optimal balance between performance and efficiency.

\textbf{Analysis of STH within STP.} 
Building on the results of ``B+IMLP,'' we further evaluated the effectiveness of the STP's subcomponent, the STH. Adding only the STH to ``B+IMLP'' yielded significant performance improvements. As shown in Tab. \ref{STH_componenet}, in the infrared-to-visible evaluation, the Rank-1 accuracy increased from 68.1\% to 69.6\% (+1.5\%), and the mAP improved from 54.6\% to 56.9\% (+2.3\%). Similarly, in the visible-to-infrared evaluation, the Rank-1 accuracy increased from 67.2\% to 70.7\% (+3.5\%), and the mAP increased from 51.0\% to 54.2\% (+3.2\%). These improvements highlight STH's role as a central hub, facilitating interframe information exchange.

\textbf{Analysis of the insertion layer of the STH.} 
To determine the optimal insertion layer for STP, we conducted experiments by setting the starting insertion layer of the STH from layers 0 to 10, as shown in Tab. \ref{inser_layer}. The results show that inserting the STH at the 9th layer yields the best performance. Specifically, in the infrared-to-visible evaluation, the Rank-1 accuracy reached 74.3\%, and the mAP was 60.2\%; in the visible-to-infrared evaluation, the Rank-1 accuracy reached 74.6\%, and the mAP was 58.6\%. 
These findings suggest that the 9th layer is the most suitable for integrating spatiotemporal information. This improvement is attributed to the hierarchical representation learning of the ViT. Early layers primarily capture low-level features such as texture and local patterns, whereas deeper layers focus on encoding high-level semantic and structural information. Inserting the STH at the 9th layer allows the module to leverage high-level information for modeling spatiotemporal dynamics without disrupting low-level feature extraction, thus achieving optimal performance.

\textbf{Analysis of STA within STP.} 
Combining STA with the STH to form VLD further enhanced performance. As shown in Tab. \ref{STH_componenet}, in the infrared-to-visible evaluation, the Rank-1 accuracy increased to 74.3\% (+4.7\%), and the mAP increased to 60.2\% (+3.3\%). Similarly, in the visible-to-infrared evaluation, the Rank-1 accuracy reached 74.6\% (+3.9\%), and the mAP improved to 58.6\% (+4.4\%). This improvement is attributed to STA's ability to strengthen STH's spatiotemporal aggregation, enabling better aggregation and diffusion of spatiotemporal information.

\begin{table*}[!ht]\small
 \centering {\caption{Ablation studies on prompt template design, where $[\text{X}]_1$ $[\text{X}]_2$ ... $[\text{X}]_\text{M}$ represents learnable parameters.}\label{text_prompt}

\begin{tikzpicture}[overlay, remember picture]
    \fill[gray!30] (0.2,1.35) rectangle (17.36,0.45); % adjust coordinates to match your table cells
\end{tikzpicture}
\begin{tikzpicture}[overlay, remember picture]
    \fill[gray!10] (0.1,-0.75) rectangle (17.24,-1.12); % adjust coordinates to match your table cells
\end{tikzpicture}
\begin{tabular}{m{0.7cm}<{\centering}m{10cm}<{\centering}m{0.7cm}<{\centering}m{0.7cm}<{\centering}m{0.7cm}<{\centering}m{0.7cm}<{\centering}}
\toprule[0.8pt]
% \begin{tikzpicture}[overlay, remember picture]
%     \fill[gray!30] (-4.105,-0.6) rectangle (11.95,0.31); % adjust coordinates to match your table cells
% \end{tikzpicture}
  \multirow{2}*{Index} &\multirow{2}*{Text Prompt Template}  & \multicolumn{2}{c}{\textit{Infrared-to-Visible}}   & \multicolumn{2}{c}{\textit{Visible-to-Infrared}} \\ 
% \cline{2-5} 
\cmidrule(lr){3-4}  \cmidrule(lr){5-6}
% \cline{6-7} \cline{8-9}
 &     & R@1    & mAP & R@1    & mAP \\ 
\toprule[0.8pt]
% 1&``A photo of a X X X X person.''       & 74.2  & 60.6  &73.4    &58.9\\
% 2&``A video of a X X X X person.''    & \textbf{74.3}  & \textbf{60.2}  &\textbf{74.6}    &\textbf{58.6}\\
1&``$[\text{X}]_1$ $[\text{X}]_2$ ... $[\text{X}]_\text{M}$ person."        & 74.0  & 60.3  &74.0    &58.5\\ 
2&``A $[\text{X}]_1$ $[\text{X}]_2$ ... $[\text{X}]_\text{M}$ person."        & 74.0  & 60.3  &74.1    &58.6\\

3 &``A $[\text{X}]_1$ $[\text{X}]_2$ ... $[\text{X}]_\text{M}$ person observed in RGB and infrared sequences.''       & 73.7  & 60.0  &74.0    &58.1\\
4 &``A $[\text{X}]_1$ $[\text{X}]_2$ ... $[\text{X}]_\text{M}$ person observed in both day and night conditions.''       & \textbf{74.4}  &\textbf{60.6}  &\textbf{74.6}    &\textbf{58.6}\\

\toprule[0.8pt]
\end{tabular}}

\end{table*}

\begin{table}[!t]\centering\small
\caption{Cross-dataset evaluation results under HITSZ-VCM (VCM) $\rightarrow$ BUPTCampus (BUPT) and BUPT $\rightarrow$ VCM settings.}
\label{gen_ablity}
\begin{tabular}{m{1.5cm}<{\centering}m{0.4cm}<{\centering}m{0.4cm}<{\centering}m{0.4cm}<{\centering}m{0.4cm}<{\centering}m{0.4cm}<{\centering}m{0.4cm}<{\centering}m{0.4cm}<{\centering}m{0.5cm}<{\centering}}
\toprule[0.8pt]
\begin{tikzpicture}[overlay, remember picture]
    \fill[gray!30] (-0.3,-1.13) rectangle (8.3,0.31); % adjust coordinates to match your table cells
\end{tikzpicture}
 \begin{tikzpicture}[overlay, remember picture]
    \fill[gray!10] (-0.395,-2.3) rectangle (8.2,-2.68); % adjust coordinates to match your table cells
\end{tikzpicture}
\multirow{3}*{Methods} 
& \multicolumn{4}{c}{\textbf{VCM $\rightarrow$ BUPT}} 
& \multicolumn{4}{c}{\textbf{BUPT $\rightarrow$ VCM}} 
\\[-0.0em] 
 \cmidrule(lr){2-5} \cmidrule(lr){6-9}
& \multicolumn{2}{c}{\textit{IR-to-Vis}} & \multicolumn{2}{c}{\textit{Vis-to-IR}}  
& \multicolumn{2}{c}{\textit{IR-to-Vis}} & \multicolumn{2}{c}{\textit{Vis-to-IR}}  
\\ \cmidrule(lr){2-3} \cmidrule(lr){4-5} \cmidrule(lr){6-7} \cmidrule(lr){8-9} 

& {R@1} & mAP & {R@1} & mAP & {R@1} & mAP & {R@1} & mAP \\ \toprule[0.8pt]
CLIP-ReID   &10.0 &11.0 &11.3 &14.3 &35.0 &23.0 &38.9 &22.8\\
TF-CLIP     &13.6 &15.5 &13.5 &16.3 &33.6 &21.7 &34.8 &19.5\\
 \hline
B     &10.9 &14.5 &17.0 &19.1 &33.2 &21.6 &35.1 &19.5\\
\textbf{our}  &\textbf{19.2} &\textbf{21.1} &\textbf{25.2} &\textbf{27.4}  &\textbf{40.8} &\textbf{28.7} &\textbf{44.8} &\textbf{27.0}\\

\toprule[0.8pt]
\end{tabular}
\end{table}

\textbf{\textit{3) Prompt design.}}  
Tab. \ref{text_prompt} shows the ablation study results on the text prompt template design for the proposed VLD. We tested different templates to analyze their impact on performance. Simple templates, such as ``$[\text{X}]_1$ $[\text{X}]_2$ ... $[\text{X}]_\text{M}$ person" (Template 1) and ``A $[\text{X}]_1$ $[\text{X}]_2$ ... $[\text{X}]_\text{M}$ person" (Template 2), deliver stable baseline performance, achieving a Rank-1 accuracy of 74.0\% and a mAP of 60.3\% in the infrared-to-visible evaluation and a Rank-1 accuracy of 74.0\%-74.1\% and mAP of 60.3\%-60.4\% in the visible-to-infrared evaluation.  
In contrast, complex descriptions, such as ``A $[\text{X}]_1$ $[\text{X}]_2$ ... $[\text{X}]_\text{M}$ person observed in RGB and infrared sequences" (Template 3), result in a slight performance drop, possibly due to CLIP's difficulty in interpreting specialized terms such as RGB and infrared. However, Template 4, which emphasizes scene diversity (``A $[\text{X}]_1$ $[\text{X}]_2$ ... $[\text{X}]_\text{M}$ person observed in both day and night conditions"), achieves the best performance. It reached the Rank-1 accuracy of 74.4\% and the mAP of 60.6\% in the infrared-to-visible evaluation and the Rank-1 accuracy of 74.6\% and the mAP of 58.6\% in the visible-to-infrared evaluation.  
These results indicate that incorporating scene diversity, such as day and night conditions, significantly enhances cross-modal matching performance.

\textbf{\textit{4) Cross-dataset generalization of VLD.}} 
To evaluate the generalizability of our method under domain and modality shifts, we conducted cross-dataset evaluations between the HITSZ-VCM (VCM) and BUPTCampus (BUPT) datasets. Specifically, the model is trained on one dataset and directly tested on the other without any fine-tuning. The corresponding results are reported in Table~\ref{gen_ablity}.
Compared with the baseline (denoted as B), our proposed method (denoted as ours) consistently achieves substantial improvements under all evaluation protocols. Under the VCM $\rightarrow$ BUPT setting, our method achieves gains of +8.3\% in Rank-1 and +6.6\% in mAP for the infrared-to-visible retrieval and +8.2\% in Rank-1 and +8.3\% in mAP for the visible-to-infrared retrieval. Similarly, under the BUPT $\rightarrow$ VCM setting, improvements of +7.6\% Rank-1 / +7.1\% mAP and +9.7\% Rank-1 / +7.5\% mAP are observed for the respective retrieval tasks.
In addition, when compared with CLIP-based competitors, including CLIP-ReID and TF-CLIP, our method exhibits superior performance across all evaluation protocols.
These results demonstrate that the proposed STP and IMLP modules substantially improve the model's robustness against domain and modality shifts. Specifically, STP captures temporal dependencies to reinforce sequence-level feature consistency, while IMLP provides modality-invariant identity supervision through language-guided alignment. The observed gains confirm that our method effectively unlocks the potential of CLIP for learning transferable identity representations in the VVI-ReID task.

\begin{table}[!t]\centering\small
\caption{Comparison of the training time and achieved performance, where IR represents infrared and VIS represents visible.}
\label{computong_cost_traing}
\begin{tabular}{m{2.2cm}<{\centering}m{1.6cm}<{\centering}m{0.5cm}<{\centering}m{0.5cm}<{\centering}m{0.5cm}<{\centering}m{0.5cm}<{\centering}}
\toprule[0.8pt]
% \rowcolor[HTML]{f5e9fb}
\begin{tikzpicture}[overlay, remember picture]
    \fill[gray!30] (-0.66,-0.6) rectangle (7.67,0.31); % adjust coordinates to match your table cells
\end{tikzpicture}
 \begin{tikzpicture}[overlay, remember picture]
    \fill[gray!10] (-0.76,-1.81) rectangle (7.58,-2.14); % adjust coordinates to match your table cells
\end{tikzpicture}
\multirow{2}*{Methods}  & \multirow{2}*{Training time}  & \multicolumn{2}{c}{\textit{IR-to-Vis}} & \multicolumn{2}{c}{\textit{Vis-to-IR}}  
\\ \cmidrule(lr){3-4} \cmidrule(lr){5-6} 
% \rowcolor[HTML]{f5e9fb}
                            &       & {R@1}    & mAP    & {R@1}   & mAP           \\ \toprule[0.8pt]

% \multirow{5}*{Image} 
 % \multirow{6}*{Video} 

 MITML\cite{lin2022learning}       &$\approx$\textbf{10 hours}     &63.7   &45.3 &64.5  &47.7   \\
   CLIP-ReID\cite{li2023clip}       &$\approx$\textbf{6.3 hours}     &58.7   &45.3 &60.4  &43.5   \\
  TF-CLIP\cite{yu2023tf}       &$\approx$\textbf{5.3 hours}     &62.3   &47.5 &62.2  &45.5   \\
 % \hline

 \textbf{our}                        &$\approx$\textbf{2 hours}      & \textbf{74.3}  & \textbf{60.2}  & \textbf{74.6}  & \textbf{58.6} \\
 \toprule[0.8pt]
\end{tabular}
\end{table}

\begin{table}[!t]\centering\small
\caption{The computational costs of the proposed method.}
\label{computong_cost_test}
\begin{tabular}{m{2.5cm}<{\centering}m{2cm}<{\centering}m{2cm}<{\centering}}
\toprule[0.8pt]
% \rowcolor[HTML]{f5e9fb}
\begin{tikzpicture}[overlay, remember picture]
    \fill[gray!30] (-0.79,-0.06) rectangle (6.96,0.32); % adjust coordinates to match your table cells
\end{tikzpicture}
\begin{tikzpicture}[overlay, remember picture]
    \fill[gray!10] (-0.91,-1.65) rectangle (6.85,-1.98); % adjust coordinates to match your table cells
\end{tikzpicture}
Methods  & Params/M  & FLOPs/G  
% \\ \cmidrule(lr){3-4} \cmidrule(lr){5-6} 
% \rowcolor[HTML]{f5e9fb}
                                     \\ \toprule[0.8pt]

% \multirow{5}*{Image} 
 % \multirow{6}*{Video} 
 % Baseline       &86.167     &83.737     \\
 % \hline
       Baseline-R50       &23.51     &8.16\\
    MITML\cite{lin2022learning}    &159.12\textcolor[rgb]{1,0,0}{(+135.61)}  &12.20\textcolor[rgb]{1,0,0}{(+4.04)}\\ \hline
  Baseline-ViT       &86.17     &13.96
  % \\
      % CLIP-ReID\cite{li2023clip}       &86.17     &13.96
  \\
    TF-CLIP\cite{yu2023tf}       &104.26\textcolor[rgb]{1,0,0}{(+18.09)}     &15.16\textcolor[rgb]{1,0,0}{(+1.2)}\\

 \textbf{our} 
 &\textbf{88.56}\textcolor[rgb]{0,0.5,0}{(+2.39)}        & 
 \textbf{14.08}\textcolor[rgb]{0,0.5,0}{(+0.12)} \\
 \toprule[0.8pt]
\end{tabular}
\end{table}

\begin{figure*}[t!]
\centering
\includegraphics[width=18.5cm,keepaspectratio=true]{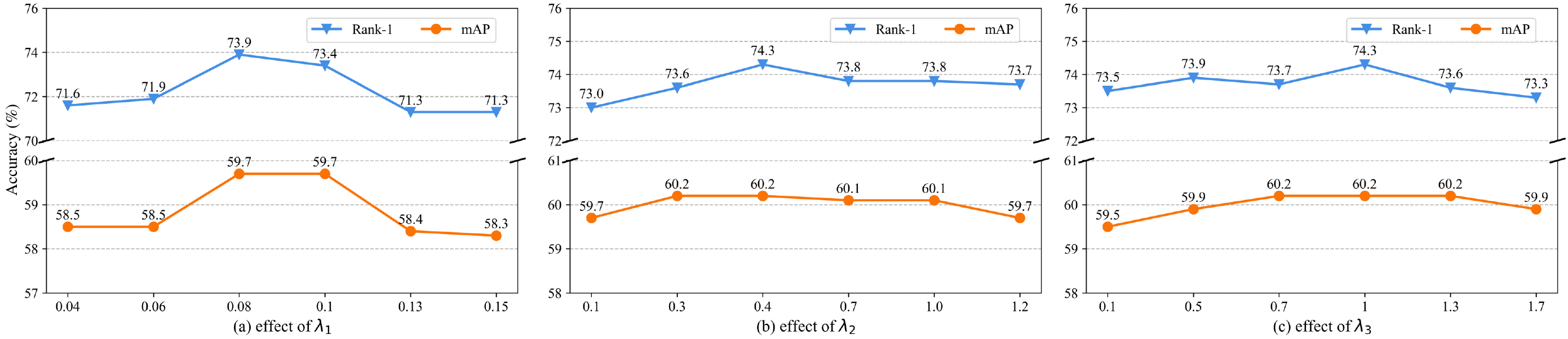}
\caption{Results of Rank-1 and mAP with different values of $\lambda_{1}$, $\lambda_{2}$ and $\lambda_{3}$ on the HITSZ-VCM dataset. }
\label{fig:param}
\end{figure*}
\textbf{\textit{5) Computational cost analysis.}} 
As shown in Tab. \ref{computong_cost_traing} and Tab. \ref{computong_cost_test}, our method has significant advantages in training efficiency, performance, and computational complexity. Compared with the MITML method, it reduces training time from approximately 10 hours to just 2 hours, an 80\% reduction, highlighting its exceptional efficiency. Compared with other CLIP-based methods, such as CLIP-ReID and TF-CLIP, it reduces training time by more than half.
In addition, compared with the Baseline method, our method slightly increases the parameter count from 86.17M to 88.56M (+2.39M) and the FLOPs from 13.96G to 14.08G (+0.12G). However, these increases are minimal and negligible.

Although the above comparisons already demonstrate the efficiency of our method from an empirical perspective, we further analyzed the theoretical complexity of the proposed modules under a per-sample setting. 
For the IMLP module, the additional cost stems mainly from the similarity-based visual-to-text alignment loss, which operates on the global [CLS] feature. Given $C$ identity-level text prompts, each dimension $D$, the similarity computation has a time complexity of $O(C \cdot D)$. Since the text encoder is frozen, it introduces no additional training overhead, reducing both memory and computational costs.
For the STP module, the introduced spatial-temporal hub (STH) is injected into the vision transformer (ViT) backbone at intermediate layers. Since each frame contributes a learnable STH token and these tokens are concatenated with patch tokens, the attention computation per layer involves $L = N + T$ tokens, leading to a complexity of $O(L^2 \cdot D) = O((N+T)^2 \cdot D)$. The aggregation step via STA introduces additional multihead attention over temporal slots, with complexity $O(T^2 \cdot D)$. As $T$ is small (e.g., 6), this cost remains minimal.
% Overall, both modules introduce modest increases in computational complexity, consistent with our empirical observations. 
Overall, our method achieves significant performance improvements with minimal computational overhead, thereby highlighting its efficiency and practicality.
% The overall parameter increase is limited to +2.39M (+2.8\%), and FLOPs grow by only +0.12G (+0.8\%), validating the scalability and deployability of the proposed method.

\begin{figure}[t!]
\centering
\includegraphics[width=8.8cm,keepaspectratio=true]{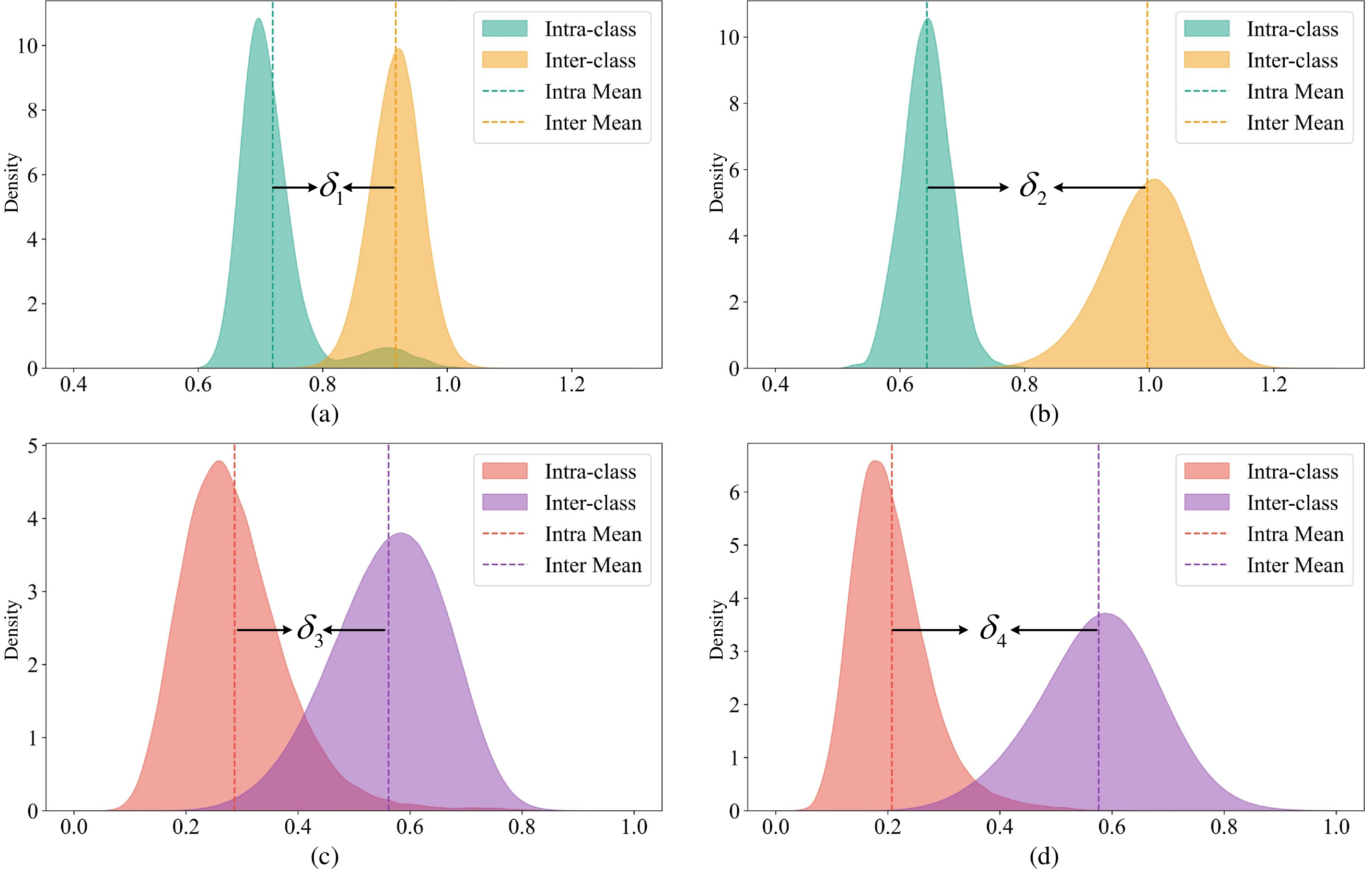}
\caption{
Visualization of intra-class and inter-class distance distributions.
(a) Distance distributions between text prompts and visual features for CLIP-ReID.
(b) Distance distributions between text prompts and visual features for VLD.
(c) Cross-modality feature distributions for CLIP-ReID.
(d) Cross-modality feature distributions for VLD.
VLD shows larger separation between intra-class and inter-class distances ($\delta_2>\delta_1$, $\delta_4>\delta_3$), demonstrating improved modality alignment and discrimination.
}
\label{dis}
\end{figure}

\subsection{Parameter Analysis}
The proposed method includes three hyperparameters, \(\lambda_1\), \(\lambda_2\), and \(\lambda_3\), which balance different loss terms. To assess their impact on performance, we conducted experiments on the infrared-to-visible evaluation mode of the HITSZ-VCM dataset, varying one parameter while keeping the others fixed.

\textbf{Impact of \(\lambda_1\):}  
In the IMLP, \(\lambda_1\) balances the alignment loss $\bm L_{v2t}$ between text and video features. This loss is key for text-video alignment, as text features reside in a lower-dimensional space than video features do. As shown in Fig. \ref{fig:param}(a), varying \(\lambda_1\) within [0.04, 0.15] reveals optimal Rank-1 accuracy and mAP at \(\lambda_1 = 0.08\). Beyond this value, performance decreases because of excessive emphasis on text features, distorting the video feature space. Thus, \(\lambda_1\) is set to 0.08 for balanced cross-modal alignment.

\textbf{Impact of \(\lambda_2\):}  
In the STH, \(\lambda_2\) controls the weight of the class-level loss $\bm L_{id}^{H}$, ensuring class-level consistency in the aggregated features. As shown in Fig.\ref{fig:param} (b), \(\lambda_2\) values within [0.1, 1.2] exhibit stable performance, peaking at \(\lambda_2 = 0.4\). This finding indicates that moderate class-level loss ensures feature consistency across modalities. Therefore, \(\lambda_2\) is set to 0.4.

\begin{figure}[t!]
\centering
\includegraphics[width=8.8cm,keepaspectratio=true]{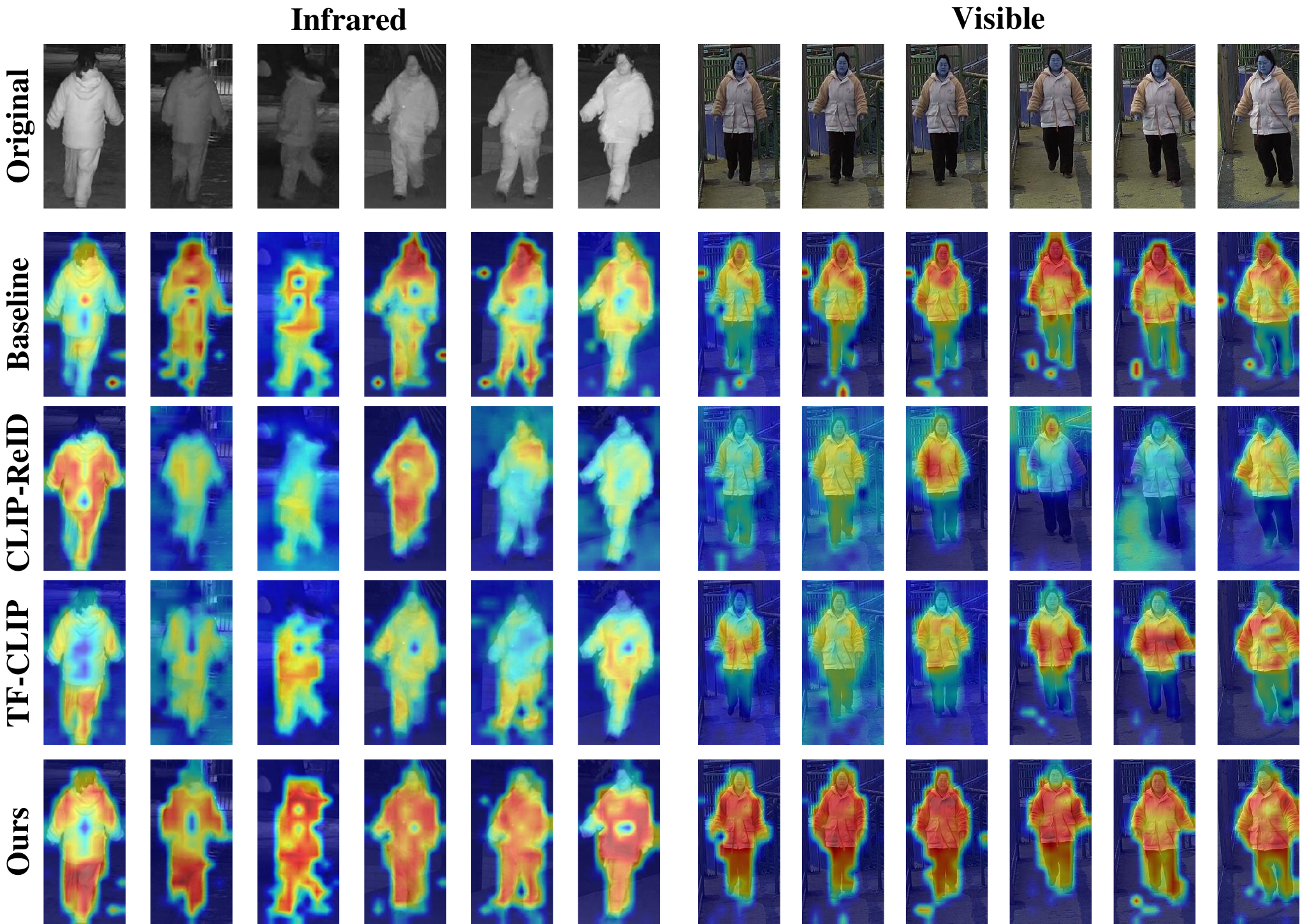}
\caption{Visualization of focus regions across different methods. We compare the proposed method VLD with Baseline, CLIP-ReID, and TF-CLIP in both the infrared and visible modalities. The infrared and visible sequences correspond to the same identity. }
\label{cam}
\end{figure}
\begin{figure*}[t!]
\centering
\includegraphics[width=17.5cm,keepaspectratio=true]{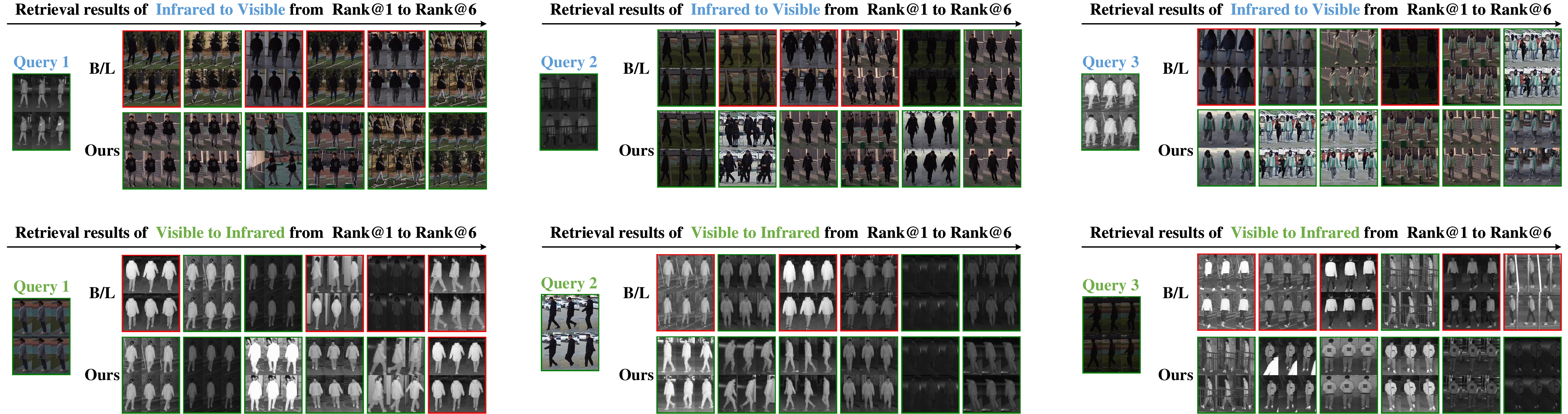}
\caption{Visualization of pedestrian search results, where B/L represents the baseline method. For each query pedestrian sequence, the top 6 most similar sequences were retrieved, with green boxes indicating correct matches (same identity) and red boxes indicating incorrect matches. }
\label{retrieval}
\end{figure*}
\textbf{Impact of \(\lambda_3\):}  
The hyperparameter \(\lambda_3\) regulates the instance-level constraint \(\bm{L}_{wrt}^{H}\) in the STH, maintaining identity consistency in aggregated spatiotemporal features. As shown in Fig.\ref{fig:param} (c), \(\lambda_3\) values within [0.1, 1.7] yield stable performance, with optimal results at \(\lambda_3 = 1.0\). Thus, \(\lambda_3\) is set to 1.0.

\subsection{Visualization}
To qualitatively validate VLD's performance, we visualized distance distributions to evaluate the effectiveness of the generated text prompts, used class activation mapping (CAM) \cite{zhou2016learning} to highlight the model's focus regions in pedestrian images, and visualized the retrieval results to show VLD's ability to retrieve pedestrian sequences effectively in complex scenarios.

\textbf{Text Prompt Analysis}  
We visualized the distance distributions between text prompts and visual features for CLIP-ReID and VLD, as well as the infrared and visible features guided by these text prompts. Fig.\ref{dis} (a) shows significant overlap in CLIP-ReID's intra-class and inter-class distances, indicating that its text prompts poorly describe pedestrian identities. In contrast, Fig.\ref{dis} (b) reveals minimal overlap in VLD, with inter-class distances significantly larger than intra-class distances, demonstrating more comprehensive identity descriptions.
This difference stems from CLIP-ReID's reliance on the original CLIP model, which lacks spatiotemporal features and struggles with infrared representation, leading to weaker visual features and difficulty in generating more comprehensive modality-shared text prompts. In contrast, VLD extracts spatiotemporal features that are modality invariant, enabling more accurate identity representation.
Fig.\ref{dis} (c) and Fig.\ref{dis} (d) further compare cross-modality feature distributions. VLD achieves smaller intra-class distances, larger inter-class distances, and minimal overlap, outperforming CLIP-ReID, which shows considerable distribution overlap. These results underscore VLD's superiority in cross-modality feature learning.

\textbf{Focus Region Analysis}  
We visualize the CAM results under visible and infrared modalities in Fig.~\ref{cam}, where warmer colors indicate stronger identity-related attention.
The baseline directly fine-tunes the CLIP visual encoder, capturing local salient cues (e.g., clothing textures) but showing a limited ability to model holistic identity structures.
CLIP-ReID \cite{li2023clip} relies on temporally unaware features from the original CLIP encoder, which lacks adaptation to VI-ReID and offers limited representation under modality discrepancies.
TF-CLIP \cite{yu2023tf} incorporates visual prototypes and temporal memory but still depends on the original CLIP encoder, leading to weak robustness under infrared inputs and attention confined to modality-specific features.
In contrast, our method augments the CLIP encoder with the STP module to capture spatiotemporal identity cues and introduces the IMLP module to ensure consistent language supervision across frames.
As shown in Fig.~\ref{cam}, our model consistently attends to identity-discriminative regions, such as the head, silhouette, and limbs, across both modalities and time, demonstrating enhanced attention consistency and semantic alignment.

\textbf{Retrieval Results Analysis}  
We visualized the retrieval performance of the baseline method and VLD in the infrared-to-visible and visible-to-infrared evaluations on the VCM-HITSZ dataset, as shown in Fig. \ref{retrieval}. 
The results demonstrate VLD's superior performance in challenging scenarios. For example, in the infrared-to-visible evaluation, Query 2 is affected by occlusion, causing the baseline method to retrieve incorrect sequences from the second result onward. In contrast, VLD accurately retrieves all sequences of the same identity. Similarly, in the visible-to-infrared evaluation, Query 3 suffers from low brightness, leading the baseline method to fail in extracting discriminative features, which results in numerous incorrect matches. However, VLD retrieves all correct sequences, thus showcasing its enhanced robustness.

\section{Conclusion}
In this paper, we propose a novel framework named VLD, which is designed to guide the model in learning modality-invariant, sequence-level pedestrian features through video-level modality-shared language prompts.
Specifically, we introduce invariant-modality language prompting (IMLP), which effectively generates modality-shared text prompts and maps visual features and text features into CLIP's multimodal space to mitigate modality differences. Additionally, we propose lightweight spatial-temporal prompting (STP), which jointly leverages the spatial-temporal hub (STH) and spatial-temporal aggregation (STA) to aggregate and diffuse spatiotemporal information across frames.
This approach enhances the integration of spatiotemporal information into text prompts, enabling the acquisition of modality-invariant sequence-level pedestrian features.
The experimental results on two publicly available VVI-ReID benchmark datasets demonstrate the significant improvements in performance and efficiency achieved by the VLD framework, thereby validating its effectiveness and superiority over other methods.

\textbf{Limitations} While the proposed VLD framework achieves superior performance, two limitations remain. First, the spatial-temporal prompting (STP) module employs a fixed-length temporal sampling strategy, which, while effective, lacks flexibility in adapting to the diverse temporal dynamics of pedestrian trajectories. Extending the model to support dynamic temporal lengths could further increase its robustness in complex video scenarios. Second, the current framework is trained and evaluated in a fully supervised setting with identity annotations. This reliance on labeled data may hinder the framework's scalability in real-world applications where annotations are limited or unavailable. Future work will investigate adaptive temporal modeling strategies and extend the proposed framework to semisupervised or unsupervised VVI-ReID paradigms.
\bibliography{ms}

\begin{thebibliography}{10}
\providecommand{\url}[1]{#1}
\csname url@samestyle\endcsname
\providecommand{\newblock}{\relax}
\providecommand{\bibinfo}[2]{#2}
\providecommand{\BIBentrySTDinterwordspacing}{\spaceskip=0pt\relax}
\providecommand{\BIBentryALTinterwordstretchfactor}{4}
\providecommand{\BIBentryALTinterwordspacing}{\spaceskip=\fontdimen2\font plus
\BIBentryALTinterwordstretchfactor\fontdimen3\font minus \fontdimen4\font\relax}
\providecommand{\BIBforeignlanguage}[2]{{%
\expandafter\ifx\csname l@#1\endcsname\relax
\typeout{** WARNING: IEEEtran.bst: No hyphenation pattern has been}%
\typeout{** loaded for the language `#1'. Using the pattern for}%
\typeout{** the default language instead.}%
\else
\language=\csname l@#1\endcsname
\fi
#2}}
\providecommand{\BIBdecl}{\relax}
\BIBdecl

\bibitem{gu2020appearance}
X.~Gu, H.~Chang, B.~Ma, H.~Zhang, and X.~Chen, ``Appearance-preserving 3d convolution for video-based person re-identification,'' in \emph{Computer Vision--ECCV 2020: 16th European Conference, Glasgow, UK, August 23--28, 2020, Proceedings, Part II 16}.\hskip 1em plus 0.5em minus 0.4em\relax Springer, 2020, pp. 228--243.

\bibitem{liu2021watching}
X.~Liu, P.~Zhang, C.~Yu, H.~Lu, and X.~Yang, ``Watching you: Global-guided reciprocal learning for video-based person re-identification,'' in \emph{Proceedings of the IEEE/CVF Conference on Computer Vision and Pattern Recognition}, 2021, pp. 13\,334--13\,343.

\bibitem{wang2021pyramid}
Y.~Wang, P.~Zhang, S.~Gao, X.~Geng, H.~Lu, and D.~Wang, ``Pyramid spatial-temporal aggregation for video-based person re-identification,'' in \emph{Proceedings of the IEEE/CVF International Conference on Computer Vision}, 2021, pp. 12\,026--12\,035.

\bibitem{bai2022salient}
S.~Bai, B.~Ma, H.~Chang, R.~Huang, and X.~Chen, ``Salient-to-broad transition for video person re-identification,'' in \emph{Proceedings of the IEEE/CVF Conference on Computer Vision and Pattern Recognition}, 2022, pp. 7339--7348.

\bibitem{liu2023deeply}
X.~Liu, C.~Yu, P.~Zhang, and H.~Lu, ``Deeply coupled convolution--transformer with spatial--temporal complementary learning for video-based person re-identification,'' \emph{IEEE Transactions on Neural Networks and Learning Systems}, 2023.

\bibitem{liu2024TMT}
X.~Liu, P.~Zhang, C.~Yu, X.~Qian, X.~Yang, and H.~Lu, ``A video is worth three views: Trigeminal transformers for video-based person re-identification,'' \emph{IEEE Transactions on Intelligent Transportation Systems}, 2024.

\bibitem{yu2023tf}
C.~Yu, X.~Liu, Y.~Wang, P.~Zhang, and H.~Lu, ``Tf-clip: Learning text-free clip for video-based person re-identification,'' in \emph{AAAI Conference on Artificial Intelligence, AAAI}, vol.~38, no.~7, 2024, pp. 6764--6772.

\bibitem{wang2022body}
Y.~Wang, G.~Qi, S.~Li, Y.~Chai, and H.~Li, ``Body part-level domain alignment for domain-adaptive person re-identification with transformer framework,'' \emph{IEEE Transactions on Information Forensics and Security}, vol.~17, pp. 3321--3334, 2022.

\bibitem{ye2021deep}
M.~Ye, J.~Shen, G.~Lin, T.~Xiang, L.~Shao, and S.~C. Hoi, ``Deep learning for person re-identification: A survey and outlook,'' \emph{IEEE transactions on Pattern Analysis and Machine Intelligence}, vol.~44, no.~6, pp. 2872--2893, 2021.

\bibitem{zheng2015partial}
W.-S. Zheng, X.~Li, T.~Xiang, S.~Liao, J.~Lai, and S.~Gong, ``Partial person re-identification,'' in \emph{Proceedings of the IEEE International Conference on Computer Vision}, 2015, pp. 4678--4686.

\bibitem{li2018harmonious}
W.~Li, X.~Zhu, and S.~Gong, ``Harmonious attention network for person re-identification,'' in \emph{Proceedings of the IEEE Conference on Computer Vision and Pattern Recognition}, 2018, pp. 2285--2294.

\bibitem{leng2019survey}
Q.~Leng, M.~Ye, and Q.~Tian, ``A survey of open-world person re-identification,'' \emph{IEEE Transactions on Circuits and Systems for Video Technology}, vol.~30, no.~4, pp. 1092--1108, 2019.

\bibitem{ye2021dynamic}
M.~Ye, C.~Chen, J.~Shen, and L.~Shao, ``Dynamic tri-level relation mining with attentive graph for visible infrared re-identification,'' \emph{IEEE Transactions on Information Forensics and Security}, vol.~17, pp. 386--398, 2021.

\bibitem{yang2023dual}
B.~Yang, J.~Chen, C.~Chen, and M.~Ye, ``Dual consistency-constrained learning for unsupervised visible-infrared person re-identification,'' \emph{IEEE Transactions on Information Forensics and Security}, 2023.

\bibitem{ye2024securereid}
M.~Ye, W.~Shen, J.~Zhang, Y.~Yang, and B.~Du, ``Securereid: Privacy-preserving anonymization for person re-identification,'' \emph{IEEE Transactions on Information Forensics and Security}, 2024.

\bibitem{ristani2016performance}
E.~Ristani, F.~Solera, R.~Zou, R.~Cucchiara, and C.~Tomasi, ``Performance measures and a data set for multi-target, multi-camera tracking,'' in \emph{European conference on computer vision}.\hskip 1em plus 0.5em minus 0.4em\relax Springer, 2016, pp. 17--35.

\bibitem{wei2018person}
L.~Wei, S.~Zhang, W.~Gao, and Q.~Tian, ``Person transfer gan to bridge domain gap for person re-identification,'' in \emph{Proceedings of the IEEE Conference on Computer Vision and Pattern Recognition}, 2018, pp. 79--88.

\bibitem{zheng2016mars}
L.~Zheng, Z.~Bie, Y.~Sun, J.~Wang, C.~Su, S.~Wang, and Q.~Tian, ``Mars: A video benchmark for large-scale person re-identification,'' in \emph{Computer Vision--ECCV 2016: 14th European Conference, Amsterdam, The Netherlands, October 11-14, 2016, Proceedings, Part VI 14}.\hskip 1em plus 0.5em minus 0.4em\relax Springer, 2016, pp. 868--884.

\bibitem{li2019GLTR}
J.~Li, J.~Wang, Q.~Tian, W.~Gao, and S.~Zhang, ``Global-local temporal representations for video person re-identification,'' in \emph{Proceedings of the IEEE/CVF International Conference on Computer Vision}, 2019, pp. 3958--3967.

\bibitem{wu2018exploit}
Y.~Wu, Y.~Lin, X.~Dong, Y.~Yan, W.~Ouyang, and Y.~Yang, ``Exploit the unknown gradually: One-shot video-based person re-identification by stepwise learning,'' in \emph{Proceedings of the IEEE Conference on Computer Vision and Pattern Recognition}, 2018, pp. 5177--5186.

\bibitem{wang2014person}
T.~Wang, S.~Gong, X.~Zhu, and S.~Wang, ``Person re-identification by video ranking,'' in \emph{Computer Vision--ECCV 2014: 13th European Conference, Zurich, Switzerland, September 6-12, 2014, Proceedings, Part IV 13}.\hskip 1em plus 0.5em minus 0.4em\relax Springer, 2014, pp. 688--703.

\bibitem{hirzer2011person}
M.~Hirzer, C.~Beleznai, P.~M. Roth, and H.~Bischof, ``Person re-identification by descriptive and discriminative classification,'' in \emph{Image Analysis: 17th Scandinavian Conference, SCIA 2011, Ystad, Sweden, May 2011. Proceedings 17}.\hskip 1em plus 0.5em minus 0.4em\relax Springer, 2011, pp. 91--102.

\bibitem{lin2022learning}
X.~Lin, J.~Li, Z.~Ma, H.~Li, S.~Li, K.~Xu, G.~Lu, and D.~Zhang, ``Learning modal-invariant and temporal-memory for video-based visible-infrared person re-identification,'' in \emph{Proceedings of the IEEE/CVF Conference on Computer Vision and Pattern Recognition}, 2022, pp. 20\,973--20\,982.

\bibitem{li2023adversarial}
H.~Li, L.~Xu, Y.~Zhang, D.~Tao, and Z.~Yu, ``Adversarial self-attack defense and spatial-temporal relation mining for visible-infrared video person re-identification,'' \emph{arXiv preprint arXiv:2307.03903}, 2023.

\bibitem{zhou2023video}
C.~Zhou, J.~Li, H.~Li, G.~Lu, Y.~Xu, and M.~Zhang, ``Video-based visible-infrared person re-identification via style disturbance defense and dual interaction,'' in \emph{Proceedings of the 31st ACM International Conference on Multimedia}, 2023, pp. 46--55.

\bibitem{du2023video}
Y.~Du, C.~Lei, Z.~Zhao, Y.~Dong, and F.~Su, ``Video-based visible-infrared person re-identification with auxiliary samples,'' \emph{IEEE Transactions on Information Forensics and Security}, vol.~19, pp. 1313--1325, 2023.

\bibitem{li2023intermediary}
H.~Li, M.~Liu, Z.~Hu, F.~Nie, and Z.~Yu, ``Intermediary-guided bidirectional spatial-temporal aggregation network for video-based visible-infrared person re-identification,'' \emph{IEEE Transactions on Circuits and Systems for Video Technology}, 2023.

\bibitem{li2023clip}
S.~Li, L.~Sun, and Q.~Li, ``Clip-reid: exploiting vision-language model for image re-identification without concrete text labels,'' in \emph{Proceedings of the AAAI Conference on Artificial Intelligence}, vol.~37, no.~1, 2023, pp. 1405--1413.

\bibitem{feng2024cross}
Y.~Feng, F.~Chen, J.~Yu, Y.~Ji, F.~Wu, T.~Liu, S.~Liu, X.-Y. Jing, and J.~Luo, ``Cross-modality spatial-temporal transformer for video-based visible-infrared person re-identification,'' \emph{IEEE Transactions on Multimedia}, 2024.

\bibitem{deng2023prompt}
C.~Deng, Q.~Chen, P.~Qin, D.~Chen, and Q.~Wu, ``Prompt switch: Efficient clip adaptation for text-video retrieval,'' in \emph{Proceedings of the IEEE/CVF International Conference on Computer Vision}, 2023, pp. 15\,648--15\,658.

\bibitem{wang2025idea}
Y.~Wang, Y.~Lv, P.~Zhang, and H.~Lu, ``Idea: Inverted text with cooperative deformable aggregation for multi-modal object re-identification,'' in \emph{Proceedings of the IEEE/CVF Conference on Computer Vision and Pattern Recognition}, 2025.

\bibitem{wang2025decoupled}
Y.~Wang, Y.~Liu, A.~Zheng, and P.~Zhang, ``Decoupled feature-based mixture of experts for multi-modal object re-identification,'' in \emph{Proceedings of the AAAI Conference on Artificial Intelligence}, vol.~39, no.~8, 2025, pp. 8141--8149.

\bibitem{wang2025mambapro}
Y.~Wang, X.~Liu, T.~Yan, Y.~Liu, A.~Zheng, P.~Zhang, and H.~Lu, ``Mambapro: Multi-modal object re-identification with mamba aggregation and synergistic prompt,'' in \emph{Proceedings of the AAAI Conference on Artificial Intelligence}, vol.~39, no.~8, 2025, pp. 8150--8158.

\bibitem{li2023logical}
S.~Li, F.~Li, J.~Li, H.~Li, B.~Zhang, D.~Tao, and X.~Gao, ``Logical relation inference and multiview information interaction for domain adaptation person re-identification,'' \emph{IEEE Transactions on Neural Networks and Learning Systems}, 2023.

\bibitem{li2020attribute}
H.~Li, Y.~Chen, D.~Tao, Z.~Yu, and G.~Qi, ``Attribute-aligned domain-invariant feature learning for unsupervised domain adaptation person re-identification,'' \emph{IEEE Transactions on Information Forensics and Security}, vol.~16, pp. 1480--1494, 2020.

\bibitem{si2022hybrid}
T.~Si, F.~He, Z.~Zhang, and Y.~Duan, ``Hybrid contrastive learning for unsupervised person re-identification,'' \emph{IEEE Transactions on Multimedia}, vol.~25, pp. 4323--4334, 2022.

\bibitem{wang2019learning}
Z.~Wang, Z.~Wang, Y.~Zheng, Y.-Y. Chuang, and S.~Satoh, ``Learning to reduce dual-level discrepancy for infrared-visible person re-identification,'' in \emph{Proceedings of the IEEE/CVF Conference on Computer Vision and Pattern Recognition}, 2019, pp. 618--626.

\bibitem{si2023tri}
T.~Si, F.~He, P.~Li, and X.~Gao, ``Tri-modality consistency optimization with heterogeneous augmented images for visible-infrared person re-identification,'' \emph{Neurocomputing}, vol. 523, pp. 170--181, 2023.

\bibitem{li2020infrared}
D.~Li, X.~Wei, X.~Hong, and Y.~Gong, ``Infrared-visible cross-modal person re-identification with an x modality,'' in \emph{Proceedings of the AAAI conference on Artificial Intelligence}, vol.~34, no.~04, 2020, pp. 4610--4617.

\bibitem{choi2020hi}
S.~Choi, S.~Lee, Y.~Kim, T.~Kim, and C.~Kim, ``Hi-cmd: Hierarchical cross-modality disentanglement for visible-infrared person re-identification,'' in \emph{Proceedings of the IEEE/CVF Conference on Computer Vision and Pattern Recognition}, 2020, pp. 10\,257--10\,266.

\bibitem{wei2021syncretic}
Z.~Wei, X.~Yang, N.~Wang, and X.~Gao, ``Syncretic modality collaborative learning for visible infrared person re-identification,'' in \emph{Proceedings of the IEEE/CVF International Conference on Computer Vision}, 2021, pp. 225--234.

\bibitem{wu2021discover}
Q.~Wu, P.~Dai, J.~Chen, C.-W. Lin, Y.~Wu, F.~Huang, B.~Zhong, and R.~Ji, ``Discover cross-modality nuances for visible-infrared person re-identification,'' in \emph{Proceedings of the IEEE/CVF Conference on Computer Vision and Pattern Recognition}, 2021, pp. 4330--4339.

\bibitem{liu2022learning}
J.~Liu, Y.~Sun, F.~Zhu, H.~Pei, Y.~Yang, and W.~Li, ``Learning memory-augmented unidirectional metrics for cross-modality person re-identification,'' in \emph{Proceedings of the IEEE/CVF Conference on Computer Vision and Pattern Recognition}, 2022, pp. 19\,366--19\,375.

\bibitem{zhang2023mrcn}
Y.~Zhang, Y.~Yan, J.~Li, and H.~Wang, ``Mrcn: A novel modality restitution and compensation network for visible-infrared person re-identification,'' \emph{arXiv preprint arXiv:2303.14626}, 2023.

\bibitem{zhang2023diverse}
Y.~Zhang and H.~Wang, ``Diverse embedding expansion network and low-light cross-modality benchmark for visible-infrared person re-identification,'' in \emph{Proceedings of the IEEE/CVF Conference on Computer Vision and Pattern Recognition}, 2023, pp. 2153--2162.

\bibitem{cheng2023cross}
D.~Cheng, X.~Wang, N.~Wang, Z.~Wang, X.~Wang, and X.~Gao, ``Cross-modality person re-identification with memory-based contrastive embedding,'' in \emph{Proceedings of the AAAI Conference on Artificial Intelligence}, vol.~37, no.~1, 2023, pp. 425--432.

\bibitem{si2023diversity}
T.~Si, F.~He, P.~Li, Y.~Song, and L.~Fan, ``Diversity feature constraint based on heterogeneous data for unsupervised person re-identification,'' \emph{Information Processing \& Management}, vol.~60, no.~3, p. 103304, 2023.

\bibitem{radford2021learning}
A.~Radford, J.~W. Kim, C.~Hallacy, A.~Ramesh, G.~Goh, S.~Agarwal, G.~Sastry, A.~Askell, P.~Mishkin, J.~Clark \emph{et~al.}, ``Learning transferable visual models from natural language supervision,'' in \emph{International Conference on Machine Learning}.\hskip 1em plus 0.5em minus 0.4em\relax PMLR, 2021, pp. 8748--8763.

\bibitem{zhou2023zegclip}
Z.~Zhou, Y.~Lei, B.~Zhang, L.~Liu, and Y.~Liu, ``Zegclip: Towards adapting clip for zero-shot semantic segmentation,'' in \emph{IEEE Conference on Computer Vision and Pattern Recognition, CVPR}, 2023, pp. 11\,175--11\,185.

\bibitem{ma2022x}
Y.~Ma, G.~Xu, X.~Sun, M.~Yan, J.~Zhang, and R.~Ji, ``X-clip: End-to-end multi-grained contrastive learning for video-text retrieval,'' in \emph{ACM International Conference on Multimedia}, 2022, pp. 638--647.

\bibitem{tang2021clip4caption}
M.~Tang, Z.~Wang, Z.~Liu, F.~Rao, D.~Li, and X.~Li, ``Clip4caption: Clip for video caption,'' in \emph{ACM International Conference on Multimedia}, 2021, pp. 4858--4862.

\bibitem{ye2025towards}
C.~Ye, Y.~Zhuge, and P.~Zhang, ``Towards open-vocabulary remote sensing image semantic segmentation,'' in \emph{Proceedings of the AAAI Conference on Artificial Intelligence}, vol.~39, no.~9, 2025, pp. 9436--9444.

\bibitem{zhu2024multi}
H.~Zhu, P.~Zhang, L.~Xue, and G.~Yuan, ``Multi-modal understanding and generation for object tracking,'' \emph{IEEE Transactions on Circuits and Systems for Video Technology}, 2024.

\bibitem{zhu2024vision}
H.~Zhu, Q.~Lu, L.~Xue, P.~Zhang, and G.~Yuan, ``Vision-language tracking with clip and interactive prompt learning,'' \emph{IEEE Transactions on Intelligent Transportation Systems}, 2024.

\bibitem{yan2023clip}
S.~Yan, N.~Dong, L.~Zhang, and J.~Tang, ``Clip-driven fine-grained text-image person re-identification,'' \emph{IEEE Transactions on Image Processing}, 2023.

\bibitem{jiang2023cross}
D.~Jiang and M.~Ye, ``Cross-modal implicit relation reasoning and aligning for text-to-image person retrieval,'' in \emph{IEEE Conference on Computer Vision and Pattern Recognition}, 2023, pp. 2787--2797.

\bibitem{he2023region}
S.~He, W.~Chen, K.~Wang, H.~Luo, F.~Wang, W.~Jiang, and H.~Ding, ``Region generation and assessment network for occluded person re-identification,'' \emph{IEEE Transactions on Information Forensics and Security}, 2023.

\bibitem{yu2025climb}
C.~Yu, X.~Liu, J.~Zhu, Y.~Wang, P.~Zhang, and H.~Lu, ``Climb-reid: A hybrid clip-mamba framework for person re-identification,'' in \emph{Proceedings of the AAAI Conference on Artificial Intelligence}, vol.~39, no.~9, 2025, pp. 9589--9597.

\bibitem{yu2024clip}
X.~Yu, N.~Dong, L.~Zhu, H.~Peng, and D.~Tao, ``Clip-driven semantic discovery network for visible-infrared person re-identification,'' \emph{arXiv preprint arXiv:2401.05806}, 2024.

\bibitem{MANet}
S.~Yan, H.~Tang, L.~Zhang, and J.~Tang, ``Image-specific information suppression and implicit local alignment for text-based person search,'' \emph{IEEE Transactions on Neural Networks and Learning Systems}, vol.~35, no.~12, pp. 17\,973--17\,986, 2024.

\bibitem{li2025breaking}
H.~Li, Y.~Liu, Y.~Zhang, J.~Li, and Z.~Yu, ``Breaking the paired sample barrier in person re-identification: Leveraging unpaired samples for domain generalization,'' \emph{IEEE Transactions on Information Forensics and Security}, 2025.

\bibitem{park2021learning}
H.~Park, S.~Lee, J.~Lee, and B.~Ham, ``Learning by aligning: Visible-infrared person re-identification using cross-modal correspondences,'' in \emph{Proceedings of the IEEE/CVF International Conference on Computer Vision}, 2021, pp. 12\,046--12\,055.

\bibitem{tian2021farewell}
X.~Tian, Z.~Zhang, S.~Lin, Y.~Qu, Y.~Xie, and L.~Ma, ``Farewell to mutual information: Variational distillation for cross-modal person re-identification,'' in \emph{Proceedings of the IEEE/CVF Conference on Computer Vision and Pattern Recognition}, 2021, pp. 1522--1531.

\bibitem{ye2021channel}
M.~Ye, W.~Ruan, B.~Du, and M.~Z. Shou, ``Channel augmented joint learning for visible-infrared recognition,'' in \emph{Proceedings of the IEEE/CVF International Conference on Computer Vision}, 2021, pp. 13\,567--13\,576.

\bibitem{feng2023shape}
J.~Feng, A.~Wu, and W.-S. Zheng, ``Shape-erased feature learning for visible-infrared person re-identification,'' in \emph{Proceedings of the IEEE/CVF Conference on Computer Vision and Pattern Recognition}, 2023, pp. 22\,752--22\,761.

\bibitem{wang2019rgb}
G.~Wang, T.~Zhang, J.~Cheng, S.~Liu, Y.~Yang, and Z.~Hou, ``Rgb-infrared cross-modality person re-identification via joint pixel and feature alignment,'' in \emph{Proceedings of the IEEE/CVF International Conference on Computer Vision}, 2019, pp. 3623--3632.

\bibitem{ye2020dynamic}
M.~Ye, J.~Shen, D.~J.~Crandall, L.~Shao, and J.~Luo, ``Dynamic dual-attentive aggregation learning for visible-infrared person re-identification,'' in \emph{Computer Vision--ECCV 2020: 16th European Conference, Glasgow, UK, August 23--28, 2020, Proceedings, Part XVII 16}.\hskip 1em plus 0.5em minus 0.4em\relax Springer, 2020, pp. 229--247.

\bibitem{zhang2021towards}
Y.~Zhang, Y.~Yan, Y.~Lu, and H.~Wang, ``Towards a unified middle modality learning for visible-infrared person re-identification,'' in \emph{Proceedings of the 29th ACM international Conference on Multimedia}, 2021, pp. 788--796.

\bibitem{yang2022learning}
M.~Yang, Z.~Huang, P.~Hu, T.~Li, J.~Lv, and X.~Peng, ``Learning with twin noisy labels for visible-infrared person re-identification,'' in \emph{Proceedings of the IEEE/CVF Conference on Computer Vision and Pattern Recognition}, 2022, pp. 14\,308--14\,317.

\bibitem{he2021transreid}
S.~He, H.~Luo, P.~Wang, F.~Wang, H.~Li, and W.~Jiang, ``Transreid: Transformer-based object re-identification,'' in \emph{Proceedings of the IEEE/CVF International Conference on Computer Vision}, 2021, pp. 15\,013--15\,022.

\bibitem{zhou2016learning}
B.~Zhou, A.~Khosla, A.~Lapedriza, A.~Oliva, and A.~Torralba, ``Learning deep features for discriminative localization,'' in \emph{Proceedings of the IEEE Conference on Computer Vision and Pattern Recognition}, 2016, pp. 2921--2929.

\end{thebibliography}

\end{document}